# *Ranking Infrared–Visible Fusion the Way Humans Do: A Learned Pairwise Preference Measure*

Haoran Liu[1,2], Mingzhe Liu[1,2,*], Peng Li[2], Guibin Zan[3,*]

1 School of Artificial Intelligence and Electronic Engineering, Sichuan Technology and Business University, Chengdu 611745, China

2 College of Nuclear Technology and Automation Engineering, Chengdu University of Technology, Chengdu 610059, China

3 National Synchrotron Radiation Laboratory, University of Science and Technology of China, Hefei 230029, China

*Corresponding author, liumz@cdut.edu.cn (Mingzhe Liu), zangb@mail.ustc.edu.cn (Guibin Zan)

## ABSTRACT

*Human pairwise comparison provides a direct basis for perceptual infrared–visible image fusion assessment, but dense annotation becomes costly as method pools grow. We present the Learned Perceptual Image Fusion Measure (LPIFM), among the earliest learned fusion assessors trained directly on dense human A/B/Tie comparisons. LPIFM jointly examines both source images and both fused candidates, combining a shared hierarchical encoder, triadic interaction, and a tie-aware objective to predict comparative preference and perceptual indifference. We construct and publicly release all 6,300 unordered comparisons among 25 methods on 21 VIFB scenes, collected through blinded, randomized annotation and expert adjudication. Across four VIFB evaluation settings, LPIFM achieves 79.2–84.0% agreement with human pairwise judgments and Spearman correlations of 0.941–0.977 with human-derived method rankings. On full method pools, accuracy exceeds the strongest of 19 conventional metrics by 16.3–21.1 pp. Consistency diagnostics show 99.98–100% candidate-swap agreement and no observed decisive preference cycles. External experiments on EVAFusion further demonstrate rapid adaptation to a different fusion-evaluation preference protocol. After only three epochs of fine-tuning, LPIFM surpasses all 19 conventional metrics in accuracy, macro-F1, and ranking correlation. LPIFM provides a scalable instrument for human-aligned fusion assessment, with the preference corpus, model weights, and code publicly available.*

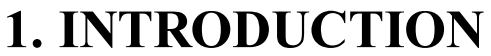

# 1. INTRODUCTION

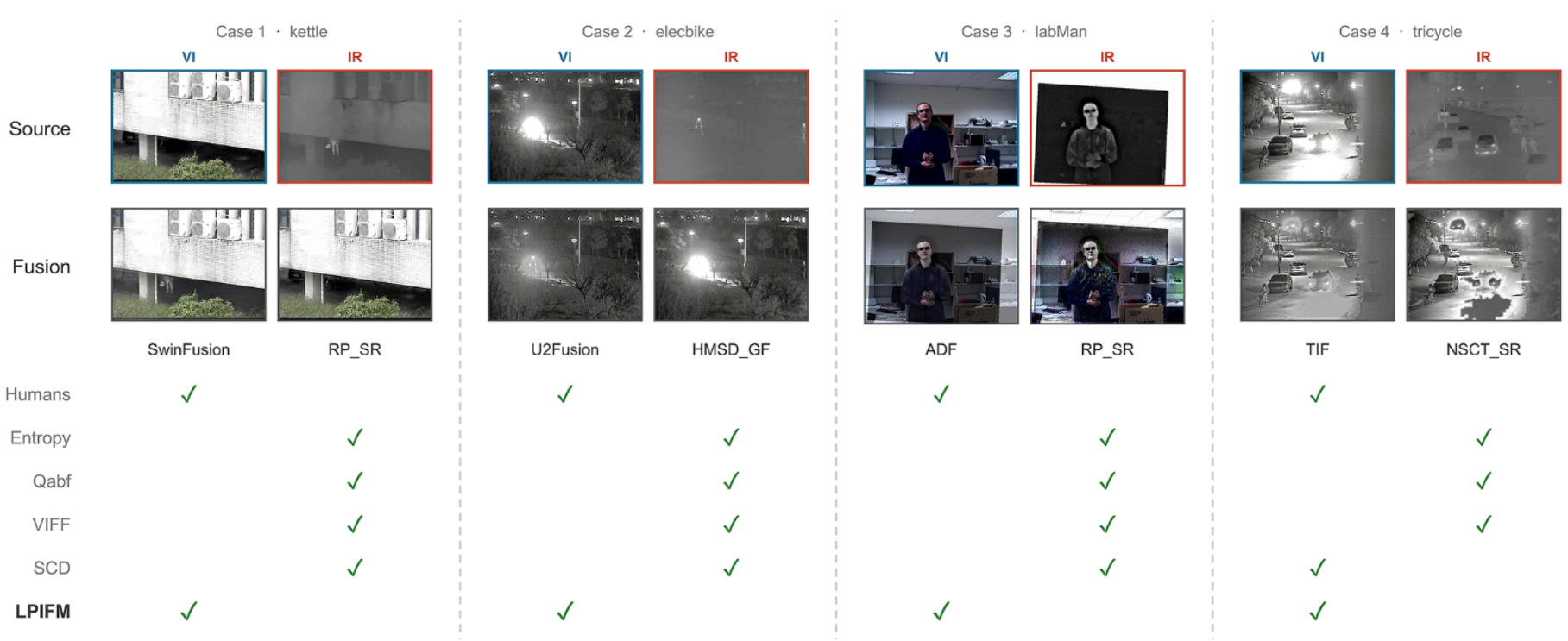

Fig. 1. Which of two fused results is preferred? Four representative comparisons from the study corpus (Case 1: kettle, SwinFusion vs. RP_SR; Case 2: elecbike, U2Fusion vs. HMSD_GF; Case 3: labMan, ADF vs. RP_SR; Case 4: tricycle, TIF vs. NSCT_SR). For each case, the visible (VI) and infrared (IR) sources are shown above the two fused candidates, and the check marks below indicate which candidate is preferred by each of the six evaluators shown (Humans, LPIFM, Entropy, Qabf, VIFF, and SCD). Human observers and LPIFM agree in all four cases, whereas the four objective metrics frequently prefer the candidate that humans reject.

Infrared–visible image fusion (IVIF) combines complementary thermal and reflectance information into a single image and now supports surveillance, driving assistance, and other downstream perception tasks. Because no ideal fused reference exists, the field ranks fusion algorithms with scalar objective metrics, and benchmark studies organize dozens of such metrics into information-theoretic, feature-based, structural-similarity, and human-perception-inspired families [1]. Each metric formalizes a different proxy for fusion quality. The VIFB benchmark explicitly notes that no single implemented metric dominates the others [1], and recent studies continue to report systematic disagreement among objective metrics on IVIF outputs [2, 3]. Fig. 1 makes the practical consequence concrete: on typical benchmark scenes, established metrics frequently prefer the fused result that human observers reject. The four cases expose two recurring failure patterns. In Cases 1 and 2, RP_SR and HMSD_GF lose salient infrared content: the pedestrian that stands out in the infrared source of the kettle scene is no longer discernible in RP_SR's output, and the rider visible in the infrared source of the elecbike scene is swallowed by amplified glare in HMSD_GF's output. Humans and LPIFM penalize this loss of infrared information, one of the decisive scoring criteria in IVIF, whereas Entropy, Qabf, VIFF, and SCD still prefer these results. In Cases 3 and 4, RP_SR and NSCT_SR introduce conspicuous artifacts (color speckle noise across the labMan scene; a dark blotch on the road and halos around the lights in the tricycle scene), and most of the same metrics reward them, because spurious structure inflates the statistics, gradients, and contrast that these proxies score as transferred information. An evaluation practice built on such proxies can therefore steer method development away from the perceptual quality it is meant to capture.

Human comparison answers the relevant question directly. Comparative subjective testing is an established methodology for relative fusion assessment and for validating objective metrics [4], and pairwise human judgments have been aggregated into method rankings with Thurstone and Bradley–Terry models [5]. Among common subjective procedures, forced-choice pairwise comparison has shown the smallest measurement variance under controlled conditions [6]. The obstacle is operational, not conceptual: evaluating $n$ algorithms requires $O(n^2)$ comparisons per scene, and every new algorithm, dataset, or ablation re-incurs the full cost of recruiting, instructing, and adjudicating observers. As algorithm pools grow, the community faces a widening gap between the evaluation protocol it trusts and the evaluation protocol it can afford to run.

Learning offers a way to remove this recurring cost, and learned IVIF assessors have begun to appear. Existing models, however, answer tasks that differ from the trusted protocol itself. Reference-based pairwise learners such as PieAPP assume a pristine reference image, which IVIF lacks [7]. Recent IVIF-specific assessors predict absolute quality scores, distill conventional metric panels, or elicit scores from multimodal language models [8-11]. None of these directly reproduces the decision that human observers actually make in the trusted protocol: a ternary A better / B better / Tie judgment about two candidates conditioned on the same infrared and visible sources. An absolute scorer must invent a scale that observers were never asked to use, and forcing a winner on perceptually indistinguishable pairs discards information that observers deliberately express.

In this paper we operationalize the human comparison protocol itself. We present the Learned Perceptual Image Fusion Measure (LPIFM), a source-conditioned model that takes the infrared source, the visible source, and two fused candidates, and predicts the same ternary decision that human observers produce. LPIFM is trained on a new dense preference corpus covering all 6,300 unordered comparisons generated by 25 fusion methods on the 21 scenes of the VIFB benchmark, labeled under a blinded, randomized, two-stage protocol with expert adjudication. A tie-aware objective preserves human indifference as a first-class outcome instead of forcing arbitrary winners. Because LPIFM outputs protocol-level decisions, its predictions plug directly into tie-aware Bradley–Terry aggregation, so a complete, human-aligned ranking of an arbitrary method pool can be computed automatically at negligible marginal cost.

Our experiments evaluate LPIFM at both the pair level and the ranking level under explicit scene- and method-generalization settings. Across four evaluation slices, LPIFM attains pairwise accuracy of 0.792–0.840 and Spearman correlation of 0.941–0.977 with human-derived rankings. Its decisive advantage appears exactly where scalable evaluation is needed most: on full 25-method pools that contain many near-ranked competitors, LPIFM maintains accuracy of 0.792–0.840 while the strongest conventional metric, SCD, falls to 0.629. This advantage has a boundary. On one six-method held-out subset whose members span nearly the full range of human-rated quality, SCD is locally competitive, and we diagnose why this strength does not transfer to realistic pools. These results indicate that trustworthy IVIF evaluation becomes affordable through direct alignment with the trusted protocol, not through better handcrafted proxies. A cross-protocol study on EVAFusion further examines whether LPIFM can adapt to a different preference convention. Its score-derived labels differ from direct human comparisons in both comparative priorities and the meaning of ties, providing a substantive explanation for the observed zero-shot disagreement. After only three epochs of fine-tuning, LPIFM achieves the highest accuracy, macro-F1, and ranking correlation among the 19 conventional metrics and LPIFM variants evaluated on the same held-out scene split. Together with the VIFB results, this demonstrates strong alignment with directly collected human preferences and rapid adaptation to the alternative preference protocol examined here.

The contributions of this work are as follows.

1. **Protocol-aligned problem formulation.** We define IVIF quality assessment as source-conditioned ternary preference prediction over two fused candidates, matching the decision space of the trusted human comparison protocol rather than an invented absolute scale.

2. **A dense human preference corpus with a full open release.** We construct a human-labeled corpus of 6,300 comparisons among 25 methods on 21 scenes, collected through blinded, randomized researcher annotation and expert adjudication. The corpus is released together with the model weights, source code, and evaluation code, so that the community gains a repeatable surrogate for human preference that can rank arbitrary IVIF method pools at scale.
3. **A tie-aware, source-conditioned comparator.** LPIFM couples a shared hierarchical encoder with a triadic interaction module and a tie-aware objective, so that human indifference is modeled explicitly rather than discarded.
4. **Systematic validation, from pair decisions to rankings to consistency.** We evaluate pairwise decisions and the induced method rankings under four scene/method generalization settings with frozen baseline calibration; we quantify the measure-like consistency of the frozen model (candidate-swap antisymmetry, preference-cycle rate, and transitivity); and we add a cross-protocol external validation on EVAFusion, in which brief fine-tuning restores leading pairwise and ranking fidelity under a different preference protocol.

## 2. RELATED WORK

### *2.1 Objective quality metrics for image fusion*

Objective fusion metrics quantify fused-image properties or source–fusion relationships through handcrafted proxies for perceptual quality. Information-theoretic metrics such as entropy [12] and mutual information [13] reward retained source statistics; feature-based metrics such as Qabf [14] reward transferred gradients; structural metrics such as SSIM [15] and MS-SSIM [16] reward preserved local structure; spatial-frequency and correlation measures follow classical image-quality practice [17]; and human-perception-inspired metrics such as Qcb [18] and Qcv [19] approximate contrast sensitivity. SCD measures the sum of correlations between source-to-fused difference images [20], FMI measures feature mutual information [21], VIFF adapts visual information fidelity to fusion [22], and Nabf quantifies fusion artifacts in the gradient-transfer family [23, 24]. These metrics are cheap and deterministic, which explains their ubiquity. However, VIFB reports that no metric dominates across scenes and methods [1], and subsequent studies document persistent disagreement among metrics and with subjective evidence [2, 3, 25]. Broader reviews of fusion methods and metrics [26-28], pseudo-reference IVIF quality evaluation [29], and registration–fusion coupling [30] further map the surrounding literature. The roots of this misalignment lie in how the scores are computed. These metrics quantify how much source signal survives into the fused image (statistics retained, gradients transferred, structure preserved) and treat every increase as a gain, regardless of whether the added content helps or harms a viewer. Noise, halos, and spurious structures raise entropy, gradient energy, and local contrast, so artifacts are routinely scored as rich information. They also aggregate uniformly over pixels, so the loss of a small but decisive region, such as a salient thermal target, barely moves a global score even though it dominates human judgment. And each fused image is scored in isolation against the sources, on a scale that is not calibrated across scenes, whereas the judgment that drives method selection is comparative. Each metric answers a question about signal preservation. None of them answers the comparative question that drives method selection, which is the question this paper targets directly.

### *2.2 Subjective comparative evaluation and preference aggregation*

Subjective testing remains the reference standard for perceptual fusion quality. Petrović established comparative subjective tests for fusion assessment and used them to validate objective metrics [4]. Loew et al. collected pairwise human judgments over fusion methods and aggregated them into rankings with Thurstone and Bradley–Terry models [5]. In the broader image-quality literature, Mantiuk et al. compared four subjective methodologies and found forced-choice pairwise comparison to yield the smallest measurement variance [6]. Indifference in paired-comparison aggregation has been treated by tie-aware Bradley–Terry models [31] and by classical analyses of ties in ranking [32]; related work on tie calibration for metric meta-evaluation provides complementary methodological guidance [33]. The strength of this line of work is validity: the protocol elicits exactly the judgment of interest. Its limitation is cost, which scales quadratically with the method pool and linearly with every new benchmark. Our work retains the protocol's decision space and evidence structure while removing the recurring human cost.

### *2.3 Learned perceptual and fusion-specific quality assessment*

Learning from human perceptual judgments is established in general image-quality assessment. PieAPP trains on pairwise preferences but assumes a pristine reference image and outputs a single-image error score [7], an assumption IVIF cannot satisfy. Neural Side-by-Side similarly learns from side-by-side human preferences, but for no-reference super-resolution evaluation rather than fusion [34]. Predictive quality models for fused long-wave infrared and visible imagery estimate subjective scores from natural-scene statistics [35]. For IVIF specifically, learned assessment has emerged only recently, and the few existing models answer questions that differ fundamentally from the one posed here. SRT applies a semantic-relation transformer that regresses an absolute quality score for a single fused image [8]; it neither conditions its judgment on a competing candidate nor admits ties, and no public implementation or trained scorer is available for re-evaluation under our protocol. EvaNet learns to approximate and consolidate conventional evaluation signals for efficient and consistent fusion assessment [9]. Its supervision targets agreement with a conventional metric panel, whereas LPIFM learns directly from human comparative judgments. Our 19-metric baseline panel evaluates the underlying conventional assessment signals; it does not constitute a numerical evaluation of EvaNet itself. EVAFusion trains an absolute reward model that grades single fused images, seeded from a small expert-annotated set and expanded with labels produced by a multimodal language model, and its primary product is preference-guided fusion generation rather than method evaluation [10]. FuScore elicits continuous quality scores from multimodal large language models with uncertainty-aware supervision built on the same label lineage [11]; both are absolute per-image scorers whose supervision passes through a language-model intermediary, and neither provides a publicly usable scorer for re-evaluation. These studies establish relevant learned approaches to fusion assessment, with differences in supervision, inference formulation, and implementation availability that affect direct comparison. We discuss those distinctions explicitly and evaluate LPIFM against 19 reproducible conventional metrics. For EVAFusion, whose scorer was unavailable for our evaluation, Section 4.8 additionally uses the released annotations to examine transfer and adaptation across preference protocols. LPIFM differs from all of them along three axes at once. It is supervised by dense, directly collected human ternary comparisons rather than metric panels, expanded labels, or language-model outputs. It performs joint two-candidate inference conditioned on both sources rather than scoring images in isolation. And it preserves ties as an explicit outcome. To our knowledge, LPIFM is among the earliest learned perceptual, fusion-specific quality assessors trained directly on human A/B/Tie pairwise comparisons, and the released dataset, model weights, and code let the community test and extend this class of assessors.

## 3. METHODS

### *3.1 Problem formulation*

Let $x_{ir}$ and $x_{vi}$ denote a registered infrared–visible source pair, and let $(y_A, y_B)$ denote two fused candidates produced from the same sources. We define IVIF quality assessment as source-conditioned ternary preference prediction: a model must map the quadruple $(x_{ir}, x_{vi}, y_A, y_B)$ to a decision in (A better, B better, Tie). This formulation matches the decision space of the human comparison protocol exactly. It requires no ideal reference, no absolute quality scale, and no assumption that all quality differences are resolvable, because the Tie outcome represents perceptual equivalence explicitly. Pair decisions over a method pool are aggregated into a method-level ranking with a tie-aware Bradley–Terry model (T-BT) [31], so the formulation connects local judgments to the algorithm-selection decisions that evaluation ultimately serves.

### *3.2 The LPIFM preference dataset*

Comparison corpus. The corpus is built on the 21 registered infrared–visible source pairs of the VIFB benchmark and fused outputs from 25 fusion algorithms [1] (representative methods in [36-49]). For each scene, observers compared every unordered pair of distinct methods, yielding $21 \times C(25,2) = 6{,}300$ human comparison units, each answered as A better, B better, or Tie. For training, every non-self comparison is represented in both candidate orders with the label mirrored, producing 12,600 ordered non-self records; 525 self-pairs labeled Tie are added, for 13,125 records in total. The ordered non-self label distribution is 5,887 A, 5,887 B, and 826 Tie. Throughout the paper we distinguish the 6,300 human comparison units from the 13,125 ordered corpus records.

Presentation and instructions. Each trial displayed four images: the visible source, the infrared source, and the two fused candidates. Candidate order was randomized and method identities were hidden. Observers judged one scene block at a time. Instructions prioritized complementary integration: retention of visible structure, texture, and readability; preservation of infrared salience and contrast; and avoidance of blur, ghosting, halo, blocking artifacts, amplified noise, unsupported structure, and false-color distortion.

Two-stage label aggregation. In the first stage, 20 bachelor-level participants provided three independent labels per comparison; majority vote yielded a preliminary label, and an IVIF researcher resolved three-way splits. This stage-1 label was retained as a reference prior, not as a vote in later aggregation. In the second stage, three IVIF researchers independently re-labeled all 6,300 unordered comparisons. The researcher labels formed the primary annotation: unanimous and two-to-one majorities were accepted by default. An additional IVIF expert then performed quality control. Three-way splits (one vote each for A better, B better, and Tie) were mandatory for adjudication and received an expert final label (129/6,300; 2.0%). Two-to-one majorities were eligible for conditional review: the expert could consult domain knowledge and the stage-1 reference and overturn the majority only when it was judged clearly unreasonable. Among the three stage-2 researchers, mean pairwise percent agreement was 76.5% and Fleiss' $\kappa$ was 0.58 (pairwise Cohen's $\kappa$ 0.54–0.60), indicating moderate inter-rater agreement with a strong majority structure. Viewing-condition records (display calibration, viewing distance, ambient illumination) were not retained and are disclosed as a reproducibility limitation in Section 5.

Dataset and code release. The annotated corpus, split manifests, model weights, and code are publicly released (see Data availability). To our knowledge, dense exhaustive pairwise human annotation at this coverage (every method pair on every scene of a public IVIF benchmark) is not otherwise publicly available.

### *3.3 Architecture*

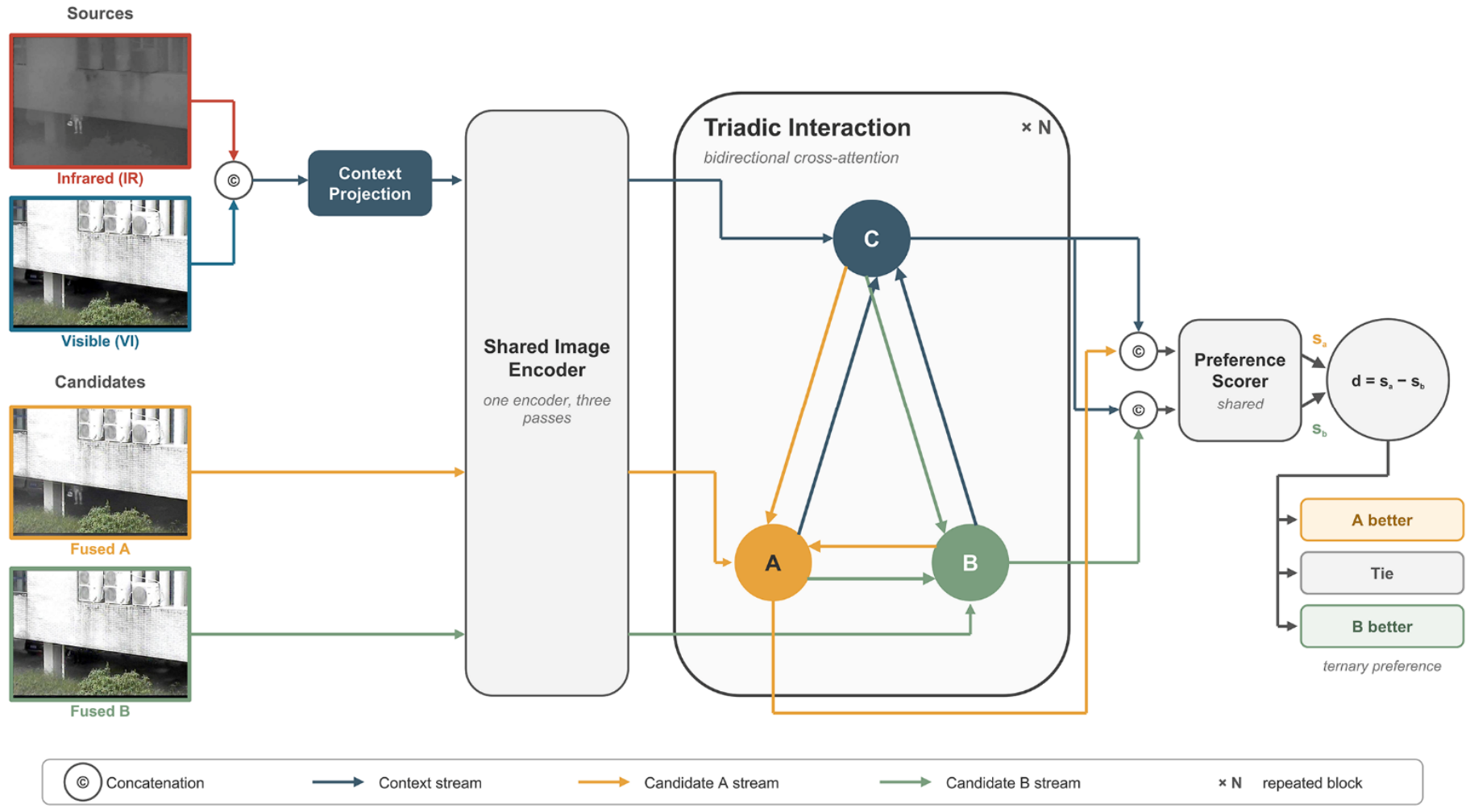

Fig. 2. LPIFM architecture. The infrared (IR) and visible (VI) sources are concatenated and passed through a context projection; the two fused candidates A and B are processed by the same shared image encoder (one encoder, three passes). The triadic interaction module applies bidirectional cross-attention (× N) among the context stream C and the candidate streams A and B. A shared preference scorer produces scores $s_A$ and $s_B$, whose difference $d = s_A - s_B$ is thresholded into the ternary decision A better / Tie / B better.

Fig. 2 shows the architecture. LPIFM has three components, each motivated by a property of the human protocol it must reproduce.

Shared image encoder with context projection. The infrared and visible sources are concatenated along the channel dimension and mapped by a learned context projection from six channels to three, yielding a source-context image that a pretrained backbone can accept. The same hierarchical encoder is then applied three times: once to this projected context and once each to the fused candidates A and B, so that all three streams share parameters and no candidate-specific weights can bias the comparison. We use a pretrained ConvNeXt V2 Base backbone at 384 × 384 input resolution [50]. The choice of a hierarchical convolutional backbone is deliberate: the backbone ablation in Section 4.5 shows a spread of more than 11 pp across eight architectures, indicating that preference learning depends on hierarchical local structure rather than on generic attention capacity.

Triadic interaction. Human observers do not judge candidates in isolation; they look back and forth between each candidate and the sources, and between the two candidates. The triadic interaction module models this behavior with bidirectional cross-attention among the context stream C and the two candidate streams A and B, stacked N times. The module lets each candidate representation encode how it preserves or distorts source information relative to both sources and relative to its competitor, which is precisely the evidence a comparative judgment needs.

Shared preference scorer. A shared scoring head maps the concatenation of the source-context representation and each interacted candidate representation to a scalar preference score, $s_A$ and $s_B$. The decision statistic is the difference $d = s_A - s_B$. Sharing the scorer across candidates, like sharing the encoder, removes any architectural asymmetry between the A and B roles.

***3.4 Tie-aware training objective***

The objective is defined on the score difference $d$ and has three terms. A preference term $L_{pref} = \text{softplus}(-y \cdot d/T)$ acts on decisive samples ($y = +1$ if A is preferred, $-1$ if B is preferred), with temperature $T$ controlling penalty sharpness. A margin term $L_{margin}$, weighted by $\lambda_m$, requires $|d|$ to reach a margin $m$ on decisive samples, so that clear human preferences map to clearly separated scores. A tie-band term $L_{tie} = \text{relu}(|d| - \tau)$, weighted by $\lambda_\tau$, acts only on tie samples and compresses $d$ into the band $[-\tau, \tau]$. The full loss is $L = L_{pref} + \lambda_m \cdot L_{margin} + \lambda_\tau \cdot L_{tie}$, with defaults $m = 1.0, \tau = 0.3, T = 1.0, \lambda_m = 0.5, \lambda_\tau = 1.0$.

This decomposition separates two questions that scalar metrics conflate: which candidate is better, and whether the pair is decidable at all. The tie band is the operative mechanism for the second question, and the ablations in Section 4.5 show that narrowing the training band to $\tau = 0.1$ reduces validation accuracy by 14.13 pp.

***3.5 Inference and ranking aggregation***

At inference, LPIFM computes the score difference $d$ for a candidate pair, calibrates it as $d_{\text{cal}} = d/T_{\text{cal}}$, and outputs A better if $d_{\text{cal}} > t$, B better if $d_{\text{cal}} < -t$, and Tie otherwise, with defaults $t = 0.3$ and $T_{\text{cal}} = 1.0$. The training preference temperature $T$ and the inference calibration temperature $T_{\text{cal}}$ are distinct; so are the model threshold $t$, the training tie band $\tau$, and the baseline calibration parameter $\kappa^*$ used to convert objective metric scores into ternary decisions. To rank a pool of $n$ methods on a scene set, LPIFM evaluates the $O(n^2)$ candidate pairs automatically and aggregates the resulting ternary decisions with tie-aware Bradley–Terry (T-BT) into a method-level strength vector and ranking. The entire pipeline is deterministic given fixed weights, so repeated evaluations of the same pool return identical rankings.

### *3.6 Implementation details*

Models were trained for 60 epochs with batch size 4, learning rate $8\times10^{-5}$ under cosine annealing with 10 warmup epochs and minimum learning rate $1\times10^{-5}$, weight decay $5\times10^{-4}$, EMA and SWA weight averaging, and global seed 200. The full model has 105.24 M parameters and 137.50 GFLOPs at 384 × 384 input. Data splits use split seed 42 with validation ratio 0.2. Authoritative run provenance (configuration and split manifests) is archived and included in the release package. The default configuration was independently trained five times (validation accuracies 77.95, 78.01, 78.73, 79.17, 80.04; mean ± s.d., 78.78 ± 0.87); the checkpoint with the highest validation accuracy (80.04%) is used for the main results, and all ablations are reported against the five-run mean.

## 4. EXPERIMENTS

### *4.1 Experimental setup*

Evaluation slices. Training labels exclude two kinds of data. First, five scenes (carLight, carShadow, manCall, running, tricycle) were held out from training updates and used for validation; we refer to them as unseen images (U-I). Because checkpoint selection used validation accuracy on these scenes, we describe U-I results as train-held-out validation generalization, not as an untouched test set. Second, six fusion methods (U2Fusion, Hybrid_MSD, MGFF, FPDE, GTF, NSCT_SR) were excluded from training labels entirely and form the unseen-method (U-M) set. The remaining 16 scenes contribute 5,776 training records; the five U-I scenes contribute 1,805 validation records. Crossing the two factors yields the four evaluation settings of Table 1.

**Table 1.** The four evaluation settings.

| Setting | Image set | Method set | Evaluation focus |
|---|---|---|---|
| U-I/U-M | Unseen images | Unseen methods | Strongest shift: scenes and algorithms both unseen |
| U-I/All-M | Unseen images | All 25 methods | Preference generalization to new |

| Setting | Image set | Method set | Evaluation focus |
|---|---|---|---|
| | | | scenes over the full pool |
| All-I/U-M | All images | Unseen methods | Isolates the unseen-algorithm factor |
| All-I/All-M | All images | All 25 methods | Overall fit on the complete corpus |

*Note: U = unseen, All = all; I = images, M = methods. All-I settings include training scenes and therefore measure overall fit rather than scene-level generalization.*

**Baselines and frozen calibration.** The conventional controls are 19 objective metrics: the 13 metrics published with VIFB [1] plus six recalculated under this protocol (VIFF [22], Nabf [23, 24], MS-SSIM [16], SCD [20], FMI [21], CC [17]). Because objective metrics output continuous scores, their pairwise score differences must be converted to ternary decisions. For each scene we compute the median absolute difference (MAD) of method-pair scores and use $\tau_{obj} = \kappa \cdot MAD$ as the tie band. We freeze $\kappa$ at $\kappa^* = 0.09$, selected once on the training seen-scene × seen-method split by matching the objective tie rate to the ground-truth tie rate (relative error 0.0163) and then frozen across all four settings. Re-selecting $\kappa$ per test setting is a test-informed oracle and is prohibited in the main results. LPIFM does not use $\kappa$; it uses its own fixed threshold $t = 0.3$. The full calibration grid is provided as supplementary Table S5. Section 2.3 discusses the supervision, task formulation, and implementation availability of related learned assessors, while Section 4.8 complements the conventional-metric comparisons with external evaluation on EVAFusion's released preference annotations.

**Outcome measures.** At the pair level we report three-class accuracy, per-class F1 ($F1_A$, $F1_B$, $F1_{Tie}$), and macro-$F1$. $F1_{Tie}$ and macro-$F1$ are reported only where ground-truth tie support is adequate (at least 30): the U-M slices contain only 2/75 and 13/315 ties, so these columns are omitted there and should be interpreted on the All-M slices (93/1,500 and 413/6,300 ties). At the method level we report Spearman $\rho$ and Kendall $\tau_b$ between predicted and human T-BT rankings, along with rank tables, mean absolute rank difference (MARD), and top-k overlap. Robustness of the ranking conclusions to the aggregation rule (top-share Ti, normalized win rate T-NR, normalized win share T-NW) is reported in supplementary Table S4.

## 4.2 Pairwise decision fidelity

Table 2 reports the full-pool comparison on scenes never used for training updates (U-I/All-M), the operating condition closest to practical use. Full results for All-I/All-M are provided in supplementary Table S6, and the cross-setting summary is retained in Table 3.

**Table 2.** Pairwise classification and T-BT rank correlation on unseen images × all methods (U-I/All-M).

| Method | Acc↑ | $F1_A$↑ | $F1_B$↑ | $F1_{Tie}$↑ | macro -$F1$↑ | $\rho$(T-BT)↑ | $\tau$(T-BT)↑ |
|---|---|---|---|---|---|---|---|
| Avg_gradient | 0.454 | 0.460 | 0.486 | 0.189 | 0.378 | -0.014 | 0.010 |
| CC | 0.568 | 0.538 | 0.615 | 0.394 | 0.516 | 0.408 | 0.311 |
| Cross_entropy | 0.478 | 0.503 | 0.482 | 0.221 | 0.402 | 0.128 | 0.100 |
| Edge_intensity | 0.458 | 0.463 | 0.489 | 0.203 | 0.385 | 0.021 | 0.040 |
| Entropy | 0.544 | 0.553 | 0.556 | 0.385 | 0.498 | 0.193 | 0.120 |
| FMI | 0.536 | 0.582 | 0.544 | 0.089 | 0.405 | 0.343 | 0.224 |
| MS-SSIM | 0.548 | 0.526 | 0.600 | 0.276 | 0.467 | 0.420 | 0.301 |
| Mutinf | 0.521 | 0.532 | 0.538 | 0.305 | 0.458 | 0.061 | 0.074 |
| Nabf | 0.489 | 0.511 | 0.500 | 0.233 | 0.415 | 0.100 | 0.080 |
| Psnr | 0.608 | 0.623 | 0.612 | 0.481 | 0.572 | 0.548 | 0.333 |
| Qabf | 0.587 | 0.611 | 0.618 | 0.099 | 0.443 | 0.430 | 0.277 |
| Qcb | 0.570 | 0.601 | 0.590 | 0.133 | 0.441 | 0.508 | 0.357 |
| Qcv | 0.604 | 0.593 | 0.660 | 0.245 | 0.499 | 0.472 | 0.341 |
| Rmse | 0.607 | 0.621 | 0.612 | 0.476 | 0.570 | 0.531 | 0.324 |
| SCD | 0.629 | 0.586 | 0.685 | 0.439 | 0.570 | 0.554 | 0.411 |
| Spatial_frequency | 0.449 | 0.469 | 0.474 | 0.101 | 0.348 | -0.040 | -0.033 |
| Ssim | 0.593 | 0.596 | 0.639 | 0.196 | 0.477 | 0.446 | 0.290 |
| Variance | 0.511 | 0.492 | 0.540 | 0.431 | 0.488 | 0.093 | 0.070 |
| VIFF | 0.577 | 0.579 | 0.617 | 0.257 | 0.484 | 0.378 | 0.250 |
| LPIFM | 0.792 | 0.820 | 0.824 | 0.215 | 0.620 | 0.941 | 0.830 |

*Note: frozen $\kappa^*$ = 0.09, calibrated on the training seen×seen split ($r_{tie}(\kappa^*)$ = 0.0624 vs. ground-truth 0.0614, relative error 0.0163). Acc is three-class accuracy including ties; $\rho/\tau$ are Spearman/Kendall $\tau_b$ against the human T-BT ranking. Ground-truth ties in this setting: 93/1,500. Metric names follow the VIFB naming convention [1].*

On training-held-out validation scenes with the full 25-method pool (Table 2), LPIFM achieved an accuracy of 0.792, 16.3 pp above SCD at 0.629. Across all images and methods (Supplementary Table S6), its accuracy was 0.840, a 21.1 pp advantage over SCD's 0.629. Macro-F1 shows the same class-averaged advantage: LPIFM reached 0.620 and 0.667 in the two settings, compared with the strongest conventional values of 0.572 and 0.536. The principal gain across both settings lies in decisive preference prediction. Tie recognition remains limited, with LPIFM $F1_{Tie}$ values of

0.215 and 0.268.

Table 3 summarizes all four settings, including the two U-M slices, and shows both halves of the evidence. LPIFM is stable everywhere: accuracy stays within 0.792–0.840 and $\rho$(T-BT) within 0.941–0.977. The conventional side is not stable. SCD is genuinely strong on the six-method held-out slices (0.840 on U-I/U-M, where it exceeds LPIFM's 0.800; 0.803 on All-I/U-M, where LPIFM reaches 0.813), yet the same metric falls to 0.629 on both full-pool settings. Section 4.4 diagnoses this contrast. Here we note only that a metric whose reliability depends on which methods happen to be compared cannot serve as a general evaluation instrument, and that full method pools are the settings practitioners face most often.

**Table 3.** Cross-setting comparison of LPIFM against the best conventional baseline.

| Setting | LPIFM Acc | LPIFM mF1 | LPIFM $\rho$ | LPIFM $\tau$ | Best Acc | Best mF1 | Best $\rho$ |
|---|---|---|---|---|---|---|---|
| U-I/U-M | 0.800 | — | 0.943 | 0.867 | 0.840 | — | 0.943 |
| U-I/All-M | 0.792 | 0.620 | 0.941 | 0.830 | 0.629 | 0.572 | 0.554 |
| All-I/U-M | 0.813 | — | 0.943 | 0.867 | 0.803 | — | 0.943 |
| All-I/All-M | 0.840 | 0.667 | 0.977 | 0.900 | 0.629 | 0.536 | 0.717 |

*Note: Best Acc / Best mF1 / Best ρ are the per-column maxima over the 19 objective metrics (excluding LPIFM); the best metric is SCD in all Acc and ρ cells. mF1 = macro-*$F1$*. Ground-truth tie support on the U-M slices (2/75 and 13/315) is below the reporting threshold of 30, so* $F1_{Tie}$ *and macro-*$F1$ *are omitted there; per-setting full tables are provided as supplementary Tables S1 and S2.*

### *4.3 Ranking fidelity*

Supplementary Table S7 provides a method-level audit of the U-I/All-M ranking. LPIFM attains a mean absolute rank difference of 1.68, compared with 5.20 for SCD, and preserves all 10 human top-10 methods; SCD preserves 7 of 10. Both LPIFM and SCD identify the human winner, U2Fusion, as rank 1. LPIFM more faithfully preserves ordering across the method pool. Other conventional metrics illustrate the broader discrepancy: FMI and Qcb rank LP_SR first although humans rank it 9th, Qcv ranks CNN first although humans rank it 7th, and Psnr ranks ADF first although humans rank it 14th.

Table 4 summarizes rank proximity across all four settings. LPIFM's MARD is lowest or tied-lowest in every setting where the pool is realistic, its top-1 agrees with humans in three of four settings, and its top-3 overlap never falls below 2/3. In the one setting without top-1 agreement (All-I/All-M), the human top two (Hybrid_MSD, U2Fusion) and LPIFM's top two (U2Fusion, Hybrid_MSD) are the same pair transposed, and the top-3 sets are identical. Replacing T-BT with win-rate aggregations (T-NR, T-NW) leaves these conclusions unchanged (LPIFM $\rho$(T-NR) = 0.940–0.976 across settings; supplementary Table S4), which indicates the ranking fidelity is a property of the predicted decisions, not of one aggregation rule.

**Table 4.** Rank proximity to the human ranking across the four settings.

| Setting | #Methods | MARD (LPIFM) | Top-1 match | Top-3 overlap | Best objective MARD | Humans top-3 | LPIFM top-3 |
|---|---|---|---|---|---|---|---|
| U-I/U-M | 6 | 0.33 | Yes | 3/3 | SCD (0.00) | U2Fusion > Hybrid_MSD > MGFF | U2Fusion > Hybrid_MSD > MGFF |
| U-I/All-M | 25 | 1.68 | Yes | 2/3 | SCD (5.20) | U2Fusion > MGFF > TIF | U2Fusion > Hybrid_MSD > TIF |
| All-I/U-M | 6 | 0.33 | Yes | 3/3 | SCD (0.33) | U2Fusion > Hybrid_MSD > MGFF | U2Fusion > Hybrid_MSD > MGFF |
| All-I/All-M | 25 | 1.20 | No | 3/3 | SCD (3.68) | Hybrid_MSD > U2Fusion > IFCNN | U2Fusion > Hybrid_MSD > IFCNN |

*Note: MARD = mean absolute rank difference against the human ranking (lower is better); best objective MARD is taken over the objective columns of the corresponding rank tables.*

### *4.4 Why SCD is locally strong on the held-out methods*

Table 3 contains an apparent anomaly: SCD reaches 0.840 accuracy on U-I/U-M and 0.803 on All-I/U-M, competitive with LPIFM, yet only 0.629 on the full pools. Table 5 locates the cause in the composition of the held-out subset.

**Table 5.** Position of the six held-out methods in the human preference ranking (All-I/All-M).

| Held-out method | Human rank (win rate) | Within-U-M win rate | Mean-SCD rank |
|---|---|---|---|
| Hybrid_MSD | 1 / 25 (0.822) | 0.819 | 9 / 25 |
| U2Fusion | 2 / 25 (0.815) | 0.824 | 1 / 25 |
| MGFF | 4 / 25 (0.740) | 0.681 | 7 / 25 |
| FPDE | 16 / 25 (0.492) | 0.438 | 17 / 25 |
| NSCT_SR | 23 / 25 (0.145) | 0.200 | 24 / 25 |
| GTF | 25 / 25 (0.060) | 0.038 | 25 / 25 |

*Note: human ranks derive from method-level win rates over All-I/All-M pairwise labels (ties counted as 0.5); within-U-M win rates count only pairs among the six held-out methods.*

The held-out pool spans human ranks 1, 2, 4, 16, 23, and 25 (Table 5), giving its comparisons an unusually broad quality range. A resampling analysis (Supplementary Section S1) confirms that this composition materially affects SCD's apparent performance. Across 3,000 random six-method subsets, SCD's mean accuracy is approximately 0.629, close to its full-pool accuracy, whereas its approximately 0.803 accuracy on the actual held-out subset lies near the 99th percentile. Subset win-rate span correlates with SCD accuracy at approximately 0.69, further linking evaluation outcomes to which methods enter the pool. Thus, SCD is competitive on the U-M slice, while LPIFM is stronger on the full 25-method pools, where its accuracy remains 0.792–0.840. These results show that subset composition can change conclusions about evaluator performance.

### *4.5 Ablation studies*

All ablations share the training protocol of Section 3.6 and report best validation accuracy; the default configuration is the five-run mean 78.78 ± 0.87 (%), and single-run deltas smaller than about 0.9 pp should not be read as stable effects.

**Table 6.** Effect of different backbones on LPIFM validation accuracy (8 architectures).

| Backbone | Params (M) | FLOPs (B) | Best Acc. (%) | *Δ* (pp) |
|---|---|---|---|---|
| **ConvNeXt-V2-Base†** | **105.24** | **137.50** | **78.78±0.87** | — |
| SwinV2-Base | 104.44 | 104.76 | 79.39 | +0.61 |
| CAFormer-B36 | 110.33 | 200.04 | 79.22 | +0.44 |
| ResNetV2-101×1 | 62.13 | 71.77 | 76.51 | -2.27 |
| ViT-B/16-384 | 103.11 | 119.53 | 73.52 | -5.26 |
| DeiT3-B/16-384 | 103.13 | 119.53 | 72.63 | -6.15 |
| FastViT-MA36 | 60.79 | 54.88 | 71.63 | -7.15 |
| EfficientNetV2-M | 70.93 | 47.85 | 67.65 | -11.13 |

*Note: † the default configuration; accuracy is the mean ± s.d. of five independent runs, other rows are single runs. Δ is computed against the default mean.*

Table 6 shows substantial backbone sensitivity while confirming that LPIFM remains competitive with several modern architectures. SwinV2 reaches 79.39%, 0.61 pp above the 78.78 ± 0.87% default mean, while using 104.76 GFLOPs versus the default's 137.50 GFLOPs. CAFormer reaches 79.22% (+0.44 pp) at 200.04 GFLOPs. ViT and DeiT3 are lower at 73.52% (−5.26 pp) and 72.63% (−6.15 pp). Both positive gains are smaller than the default run-to-run standard deviation; ConvNeXt-V2-Base remains the default.

**Table 7.** Ablation on the capacity of the triadic interaction module.

| ID | Config | Depth | Embed | Heads | Params (M) | Best Acc. (%) | *Δ* (pp) |
|---|---|---|---|---|---|---|---|
| E0 | None | 0 | 1024 | — | 97.15 | 77.40 | -1.38 |
| E1 | Nano | 1 | 256 | 4 | 88.85 | 75.96 | -2.82 |
| E2 | Tiny | 1 | 512 | 4 | 91.22 | 69.20 | -9.58 |
| E3 | Small | 2 | 512 | 4 | 91.94 | 78.23 | -0.55 |

| ID | Config | Depth | Embed | Heads | Params (M) | Best Acc. (%) | *Δ* (pp) |
|---|---|---|---|---|---|---|---|
| E4 | Medium | 2 | 768 | 8 | 96.45 | 75.29 | -3.49 |
| **E5†** | **Base** | **3** | **1024** | **8** | **105.24** | **78.78±0.87** | **—** |
| E6 | Large | 4 | 1024 | 8 | 107.93 | 78.73 | -0.05 |

*Note: † the default configuration (five-run mean ± s.d.); other rows are single runs. Δ is computed against the E5 mean. E0 removes the triadic interaction module while retaining the default 1024-d embedding and scorer; E1–E4 use smaller Embed widths, which also shrink the other parts of the model and therefore reduce total parameter count even when a module is present.*

Removing the triadic interaction module (E0) reduces accuracy from the 78.78% default mean to 77.40%, a 1.38 pp difference. Configuration remains consequential: Tiny reaches 69.20% (−9.58 pp), whereas Small and Large reach 78.23% and 78.73%. The narrower configurations also change other model components, so these rows compare complete configurations rather than interaction width alone. Base is retained as the default configuration.

**Table 8.** Ablation on pairwise preference loss hyperparameters.

| ID | Setting | Changed value | Best Acc. (%) | *Δ* (pp) |
|---|---|---|---|---|
| **P1†** | **Baseline** | $\boldsymbol{m = 1.0, \tau = 0.3, T = 1.0, \lambda_m = 0.5, \lambda_\tau = 1.0}$ | **78.78±0.87** | — |
| P2 | Small margin | $m = 0.5$ | 76.45 | -2.33 |
| P3 | Large margin | $m = 2.0$ | 77.17 | -1.61 |
| P4 | Narrow tie band | $\tau = 0.1$ | 64.65 | -14.13 |
| P5 | Wide tie band | $\tau = 0.7$ | 75.07 | -3.71 |
| P6 | Low temperature | $T = 0.5$ | 77.73 | -1.05 |
| P7 | High temperature | $T = 2.0$ | 72.74 | -6.04 |
| P8 | No margin loss | $\lambda_m = 0$ | 78.45 | -0.33 |
| P9 | Strong margin loss | $\lambda_m = 1.0$ | 79.17 | +0.39 |
| P10 | Weak tie loss | $\lambda_\tau = 0.5$ | 80.06 | +1.28 |

*Note: † the default configuration (five-run mean ± s.d.); other rows are single runs changing one hyperparameter each. Δ is computed against the P1 mean.*

Among the loss perturbations in Table 8, reducing the training tie-band parameter to $\tau = 0.1$ produces the largest drop, reaching 64.65% (−14.13 pp). The training parameter $\tau$ is distinct from the fixed inference threshold $t = 0.3$. A wider training band, $\tau = 0.7$, decreases accuracy by 3.71 pp, while $T = 2.0$ produces a non-marginal 6.04 pp decrease. Weak tie-loss weighting (P10) reaches 80.06%, essentially matching the best default run at 80.04%. Together, the results show that performance is especially sensitive to the narrow training tie band and high temperature, while the weaker tie-loss weight remains competitive with the default.

Supplementary Table S8 shows that OneCycle reduces accuracy by 8.70 pp and that lowering the minimum learning rate to $1\times10^{-6}$ reduces it by 15.23 pp.

### *4.6 Qualitative case analyses*

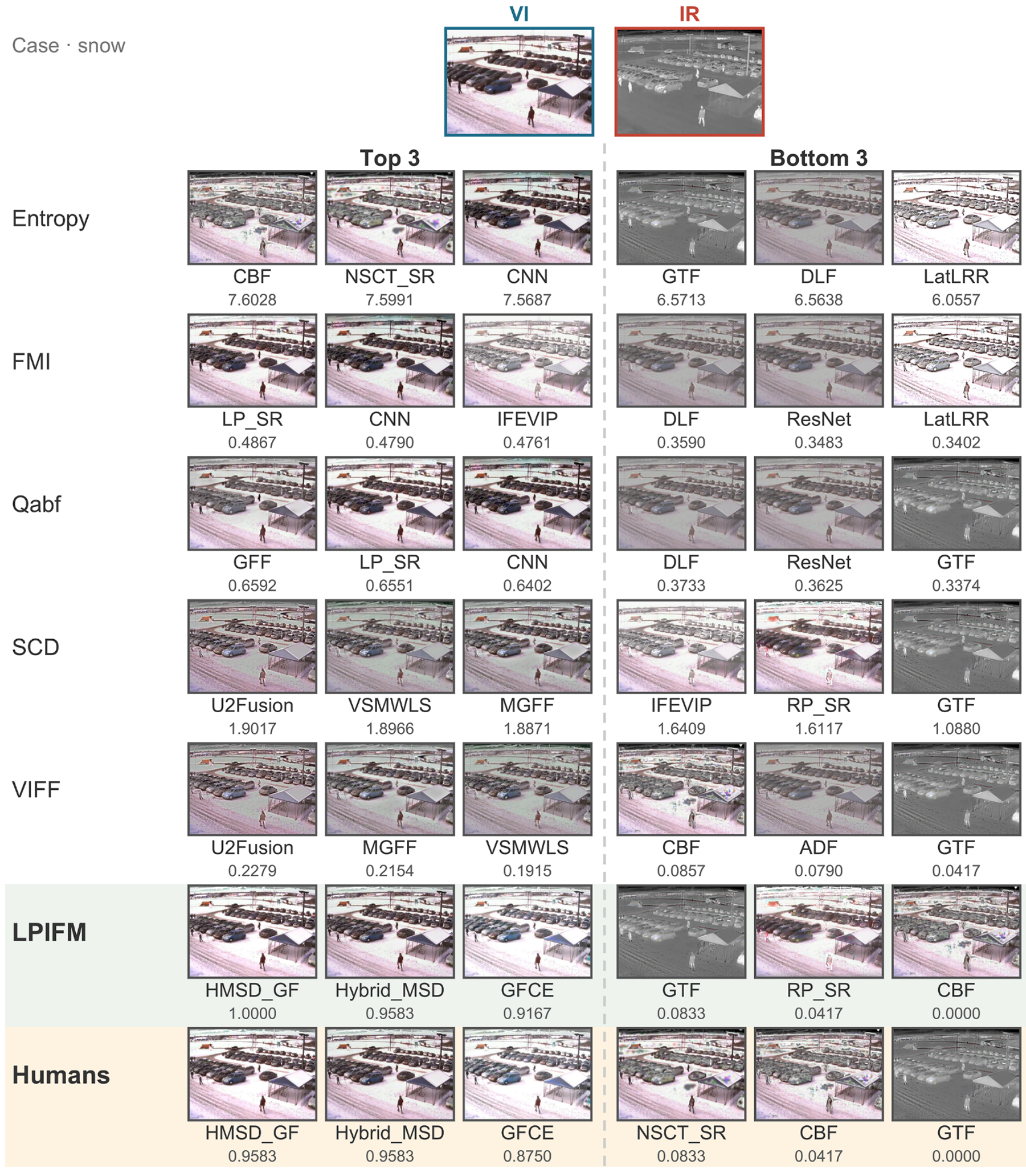

Fig. 3. Per-scene ranking case study on the snow scene. For each evaluator (rows: Entropy, FMI, Qabf, SCD, VIFF, LPIFM, Humans), the three highest-ranked and three lowest-ranked fused results are shown with the corresponding scores (classical metrics: native scalar values; LPIFM and Humans: per-scene normalized win rates). LPIFM's top three (HMSD_GF, Hybrid_MSD, GFCE) coincide with the human top three.

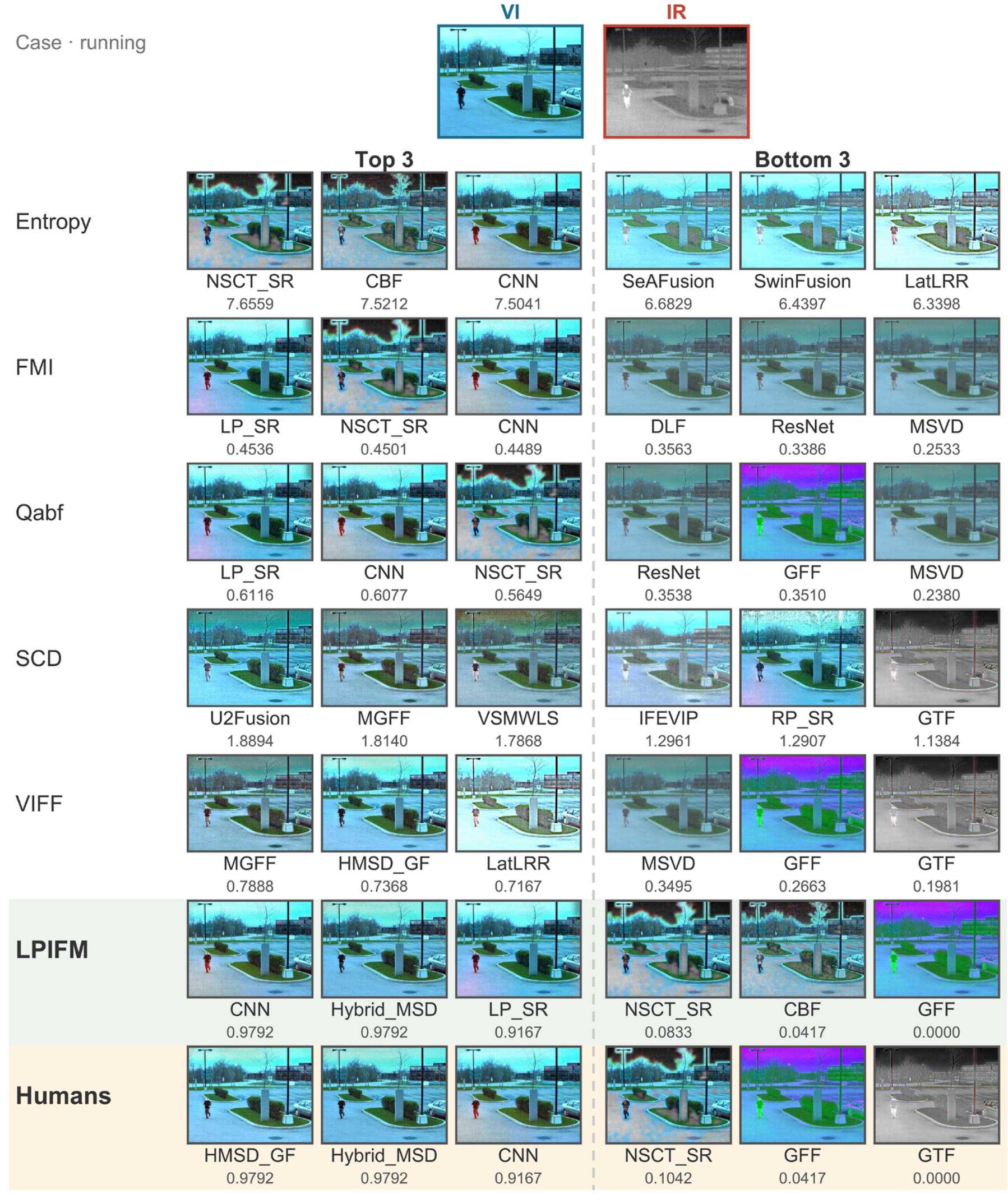

Fig. 4. Per-scene ranking case study on the running scene (a validation scene never used for training updates), in the format of Fig. 3. LPIFM's top-three set shares two methods with the human top-three set, while several classical metrics elevate methods placed near the bottom by humans.

Figs. 3 and 4 examine two representative scenes from a practitioner's perspective, focusing on which methods each evaluator ranks near the top and bottom. On the snow scene (Fig. 3), LPIFM's top three (HMSD_GF 1.0000, Hybrid_MSD 0.9583, and GFCE 0.9167) contain exactly the same methods as the human top three. On the running scene (Fig. 4), a validation scene excluded from training updates, LPIFM's leading group (CNN 0.9792, Hybrid_MSD 0.9792, and LP_SR 0.9167) shares two methods with the human group (HMSD_GF 0.9792, Hybrid_MSD 0.9792, and CNN 0.9167). In both scenes, classical metrics repeatedly rank results near the top that humans place near the bottom, consistent with the quantitative analysis in Section 4.3.

Visual inspection of Figs. 3 and 4 and Supplementary Figs. S5 and S6 clarifies the source of these disagreements. Results favored by classical metrics often contain visible defects. NSCT_SR, whose outputs contain speckle and smoke-like spurious structures, is placed in the top three by at least one of Entropy, FMI, and Qabf in each of the four scenes. On the snow and running scenes, Entropy ranks NSCT_SR and CBF among its top three even though human observers place them at or near the bottom. These artifacts increase the image statistics that the metrics interpret as information. The bottom rankings show a similar discrepancy. Humans tend to place artifact-heavy or degraded results, including NSCT_SR, CBF, GTF, and MSVD, in their bottom three, whereas classical metrics sometimes demote mid-ranked results such as DLF and ResNet (Fig. 3 and Supplementary Fig. S5). For every scene, LPIFM's top-three and bottom-three sets each share at least two methods with the corresponding human sets.

Supplementary Figs. S1-S6 extend the analysis to the kettle, elecbike, labMan, tricycle, walking, and man scenes. LPIFM and the human observers agree that SwinFusion is the best method on the kettle scene (Fig. S1), and their top-three sets are identical on the elecbike scene (Fig. S2). Fig. S4 also records an informative failure case. On the tricycle scene, LPIFM places GTF in its leading group (0.9375), whereas the human observers rank it near the bottom (0.0417). Because tricycle is a validation scene and GTF is a held-out method, this case combines both generalization factors. Its implications for scene-level reliability are discussed in Section 5.

***4.7 Measure-like consistency diagnostics***

We evaluated three consistency properties of the frozen SWA checkpoint, keeping the decision threshold $t = 0.3$ and calibration temperature $T_{cal} = 1.0$ fixed. Candidate-swap antisymmetry requires decisive verdicts to reverse, ties to remain ties, and the soft preference gap to negate when candidate order is exchanged. Because the gold labels are unordered, human left–right swap consistency cannot be assessed. On U-I/All-M, LPIFM satisfies the hard swap criterion for all 1,500 pairs; the forward gap and negated swapped gap have correlation 1.0000, with mean absolute residual 0.0083.

Cycle and transitivity eligibility are defined separately. A cycle is evaluated only for a triplet with three decisive edges and occurs when those edges form a circular preference. LPIFM has 0 cycles among 10,054 eligible triplets, compared with 70 among 9,520 human triplets, a human rate of 0.0074. Transitivity begins with each decisive two-edge path, such as A > B and B > C. The hard criterion requires a decisive closing verdict A > C; the weak criterion also accepts a tie on the closing comparison. LPIFM achieves hard and weak transitivity of 1.0000 over 10,054 paths, while the human rates are 0.9624 and 0.9786 over 9,819 paths.

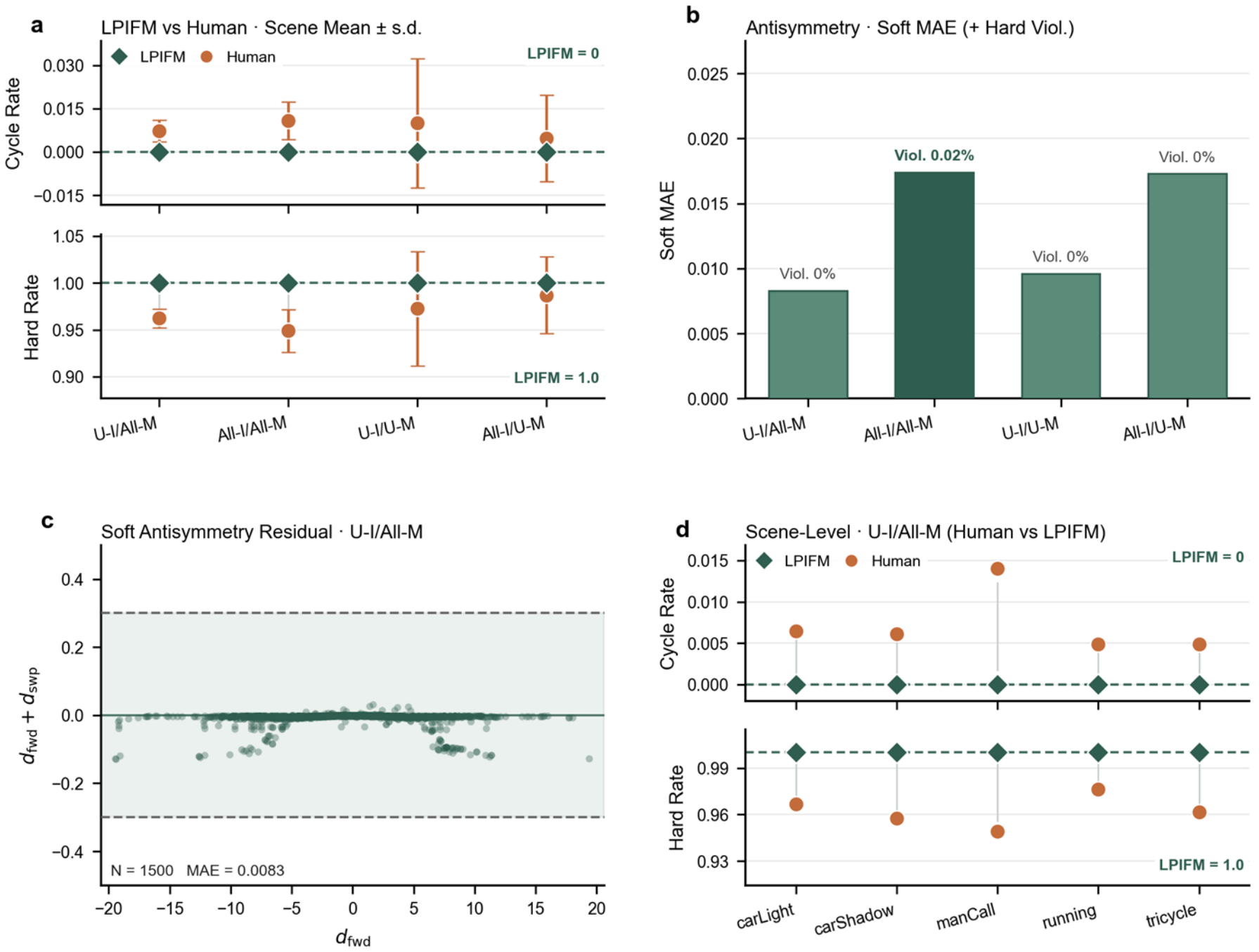

Fig. 5. Measure-like consistency of LPIFM across the four VIFB protocol splits, evaluated using the frozen checkpoint and decision threshold ($t = 0.3$). (a) Scene-level mean ± s.d. of the decisive-triplet cycle rate (top) and hard transitivity (bottom) for humans (circles) and LPIFM (diamonds). (b) Soft antisymmetry residual MAE by split; labels above the bars indicate hard swap-violation rates. (c) Forward–swap residual $d_{fwd} + d_{swp}$ plotted against $d_{fwd}$ on U-I/All-M ($N = 1{,}500$); dashed lines mark ± t. (d) Per-scene cycle and hard-transitivity rates for the five U-I/All-M scenes.

Across the other protocol splits, swap agreement is at least 99.98%, the cycle count remains zero, and hard and weak transitivity remain 1.0000. Fig. 5 summarizes these diagnostics across splits; further details are provided in Supplementary Section S2. The results support high internal consistency under the reported frozen settings and definitions. Internal consistency remains distinct from agreement with human preferences, which is measured by the pairwise and ranking results.

### *4.8 Cross-protocol external validation on EVAFusion*

The experiments above validate LPIFM against the protocol it was built to reproduce: dense human pairwise adjudication on VIFB. They leave open how the instrument behaves when both the scene pool and the preference protocol change. The obstacle to answering this question directly is the same cost asymmetry that motivates LPIFM in the first place. The VIFB preference corpus required adjudicating all 6,300 unordered comparisons under a blinded, randomized, two-stage protocol with expert quality control. Because this labeling cost grows quadratically with the number of methods, repeating a campaign of comparable density on a second benchmark was not feasible for this work. We therefore turned to the most relevant existing IVIF corpus with human-grounded preference annotation that we could identify: EVAFusion [10], whose quality scores originate from senior IVIF experts and are expanded and reviewed at scale. This section uses EVAFusion for a cross-protocol external validation with three aims: to stress-test zero-shot transfer, to diagnose the observed gap through explicit dataset and annotation-protocol contrasts, and to measure whether a brief fine-tune restores pairwise and ranking fidelity on a held-out scene split.

EVAFusion is not used here as a second human A/B/Tie benchmark, and that distinction matters for every claim that follows. The resource provides absolute per-image scores on four dimensions: Thermal Retention, Texture Preservation, Artifacts, and Sharpness. The Overall score is defined as the mean of these four scores. A seed set of 100 images scored by four senior experts was expanded by a fine-tuned multimodal language model to 9,350 images and subsequently reviewed by five IVIF researchers [10]. Expert review strengthens quality control, but this lineage is not equivalent to the dense human pairwise adjudication used to train LPIFM. For the present study we convert EVAFusion into source-conditioned ternary pairs. Among the 11 fused candidates of each scene, the candidate with the higher Overall score is labeled preferred, and exact score equality is labeled Tie. Over 850 scenes, this conversion yields 46,750 derived pairs. These derived labels are proxy ground truth, not freshly adjudicated human preference. The evaluation therefore measures agreement with an Overall-derived preference regime. It does not re-validate LPIFM against a human panel of density comparable to that of VIFB. We refer to this setting as cross-protocol external validation.

We enforce a scene-disjoint split with a fixed random seed: 612 training scenes, 68 validation scenes, and 170 held-out test scenes, with no scene appearing in more than one split. Each scene contributes all 55 pairs among its 11 candidates, giving 9,350 test pairs with a ground-truth Tie rate of 0.3075 under the exact-equality rule. Classical metrics are decoded to ternary decisions with a per-scene adaptive band of $\tau_{obj} = \kappa^* \cdot MAD$ ($\kappa^* = 0.09$) on oriented scores. LPIFM retains its fixed decode settings ($t = 0.3, T_{\mathrm{cal}} = 1.0$). For this experiment, the conventional metrics retain the VIFB-calibrated $\kappa^*$ without EVAFusion-specific recalibration, while fine-tuning updates LPIFM's weights but leaves its $t$ and $T_{\mathrm{cal}}$ decoding settings fixed. Fine-tuning initializes from the released VIFB-trained checkpoint, uses learning rate $1\times10^{-5}$ with a backbone learning-rate factor of 0.5, freezes the backbone for the first epoch, and runs for three epochs in total. We report the final checkpoint on the held-out test split.

Zero-shot LPIFM transfers poorly to this proxy protocol. On the 170-scene test split the frozen model attains accuracy 0.3656, macro-$F1$ 0.3265, and mean within-scene Spearman correlation 0.0657, ranking 16th by accuracy among the 21 assessors evaluated on the same labels (Table 9; complete results in supplementary Table S9). It predicts Tie for only 8.56% of pairs against a ground-truth Tie rate of 30.75%. The combination of weak pairwise accuracy, near-zero ranking correlation, and severe under-prediction of indifference indicates that frozen LPIFM does not reproduce the EVAFusion preference regime. Explaining this gap requires contrasting the two datasets and their annotation protocols rather than invoking a single cause.

The strongest contrast concerns label statistics and Tie semantics. The VIFB corpus (21 scenes, 25 methods, 6,300 unordered pairs) was labeled by dense, fully human, source-conditioned comparison in which a Tie denotes perceptual equivalence under joint viewing of both sources and both candidates; its non-self Tie rate is 413/6,300 = 6.56%. EVAFusion pair labels are instead converted post hoc from absolute scores, and a Tie is declared exactly when two Overall scores collide. That collision is frequent because the Overall distribution over 9,350 images occupies a narrow quantized band: scores range from 2.0 to 4.75 on a 0.25 grid with only 12 distinct values, 95.9% of images fall in [3.0, 4.25], and the three most frequent values (3.75, 4.0, and 3.0) alone cover 63.3% of images. Exact-equal Ties consequently reach 28.51% of all pairs and 30.75% of test pairs, roughly 4.7 times the VIFB human Tie rate. The equality rule is also semantically weaker than a perceptual judgment: among the 13,330 equal-Overall pairs, 1,310 (9.83%) still differ on at least one of the four quality dimensions, so an EVAFusion Tie can conceal a visible multidimensional trade-off, while any 0.25 score step forces a preference even though that step is not a human just-noticeable-difference decision.

The preference functions themselves also differ, beyond their Tie conventions. VIFB observers answered a relative, source-conditioned question: with both sources and both candidates simultaneously in view, they judged which candidate better integrated the complementary information. EVAFusion instead scores each image in isolation, without competitor context, on four absolute dimensions that are averaged into a single scalar. An averaged absolute score need not reproduce comparative priorities. Supplementary Fig. S7 illustrates this divergence on two held-out test pairs. In both cases the EVAFusion Overall scores prefer one candidate, whereas zero-shot LPIFM prefers the other. To adjudicate the disagreement, the two pairs were re-labeled following the second-stage procedure of Section 3.2, in which IVIF researchers independently compare the blinded, randomly ordered candidates alongside both sources. These human judgments concur with LPIFM in both cases. Under the proxy labels, however, such pairs are counted as zero-shot errors, so the zero-shot accuracy should not be read as a purely model-side failure.

Three epochs of light fine-tuning change the operating point. The last checkpoint reaches accuracy 0.5066, macro-$F1$ 0.4941, and mean Spearman correlation 0.4449, with a predicted Tie rate of 0.2516 that moves close to the ground-truth rate of 0.3075. Relative to zero-shot, these are gains of 14.10 pp in accuracy, +0.1676 macro-$F1$, and +0.3792 Spearman. Evaluated against nineteen classical metrics on the same held-out labels, fine-tuned LPIFM leads all three primary measures among the 21 assessors compared (Table 9; complete results in supplementary Table S9). The strongest classical competitors are Qcv (accuracy 0.4802, macro-$F1$ 0.4099, Spearman 0.3840) and Qabf (0.4801, 0.4069, 0.3922); relative to the best classical result in each column, fine-tuned LPIFM improves accuracy by 2.64 pp and macro-$F1$ by 0.0842 over Qcv, and Spearman correlation by 0.0527 over Qabf.

**Table 9.** Performance of selected assessors on the scene-disjoint EVAFusion test split (170 scenes, 9,350 pairs), with exact equality in Overall score defining a Tie.

| Rank | Assessor | Acc | macro-F1 | $\rho$ |
|---|---|---|---|---|
| 1 | LPIFM fine-tuned | 0.5066 | 0.4941 | 0.4449 |
| 2 | Qcv | 0.4802 | 0.4099 | 0.3840 |
| 3 | Qabf | 0.4801 | 0.4069 | 0.3922 |
| 8 | SCD | 0.4072 | 0.3419 | 0.1804 |
| 16 | LPIFM zero-shot | 0.3656 | 0.3265 | 0.0657 |

*Note: Ranks are based on accuracy among all 21 assessors; complete results are provided in supplementary Table S9. Metric names follow the VIFB naming convention [1].*

Taken together, the VIFB and EVAFusion results support LPIFM as a learnable preference instrument: it closely reproduces directly collected human comparisons on VIFB and adapts rapidly to EVAFusion's distinct score-derived preference conventions. After only three epochs of low-learning-rate fine-tuning, the VIFB-trained model outperforms all 19 conventional metrics in accuracy, macro-F1, and ranking correlation on the held-out EVAFusion scenes. This supports the practical adaptability of the pretrained comparator across the two evaluated preference protocols. The zero-shot results also show why preference alignment matters: differences in comparative priorities, label construction, and tie semantics can change which predictions count as correct. In the two cases examined in supplementary Fig. S7, additional human comparisons support LPIFM despite disagreement with the proxy labels. These observations identify protocol mismatch as a substantive contributor to the observed disagreement, while the experiment does not isolate its contribution from changes in scene content and label lineage. For a new application, representative preference judgments can guide whether adaptation is needed. Limitations of the proxy labels and the single split seed are discussed in Section 5.

## 5. DISCUSSION

The central finding of this work is that the human pairwise comparison protocol for IVIF, long treated as trustworthy but unaffordable, can be operationalized as a learned, repeatable instrument without surrendering its decision semantics. LPIFM answers the same ternary question that the protocol puts to human observers, and its agreement with human decisions (0.792–0.840) and human rankings ($\rho = 0.941 - 0.977$) held across scene and method generalization slices on this benchmark. We interpret this stability, together with the large remaining gap between conventional metrics and LPIFM on full method pools, as evidence that the limiting factor in current IVIF evaluation is not the sophistication of handcrafted proxies but their distance from the human decision function. Once that function is learned directly, ranking fidelity follows, and it follows robustly across aggregation rules (T-BT, T-NR, T-NW).

The ablations suggest three reasons for this margin. First, the decision space matters: the tie band incurs the largest ablation loss ($-14.13$ pp when narrowed), which indicates that agreement with human judgment depends on modeling when humans decline to judge. Second, the visual prior matters: a backbone spread of more than 11 pp shows that preference discrimination requires hierarchical local structure, consistent with the fact that fusion defects (halos, ghosting, amplified noise, lost thermal salience) are local, structured phenomena. Third, joint comparison contributes a stable margin: triadic interaction adds a consistent gain at moderate capacity, complementing the supervision and formulation that drive most of the performance. These observations indicate that the dataset and the task definition matter more than any single architectural choice.

For the community, the practical significance is an instrument, not just a benchmark result. A researcher developing a fusion method can, at negligible marginal cost, obtain the decision a dense human panel would most likely have produced: which of two variants is preferable on each scene, whether the difference is perceptible at all, and where a new method sits in a human-aligned ranking of an arbitrary pool. The same instrument can serve benchmark maintainers as a meta-evaluation layer, flagging metrics or leaderboard configurations that diverge from human preference. To support this use, we publicly release the annotated preference dataset, model weights, source code, and evaluation code (see Data availability), so that others can retrain and audit preference-aligned assessors. The release converts a single trained model into a reproducible research direction.

Rival explanations for the headline results deserve explicit treatment. The first is that a strong handcrafted metric might suffice. Section 4.4 traces SCD's local parity to the pairing of an extreme-span held-out subset with SCD's strength on coarse contests: its accuracy on this subset lies near the 99th percentile of random six-method subsets, whereas its mean accuracy across those random subsets is approximately 0.629, matching its full-pool value. The second is benchmark-specific fitting. The U-I slices provide scene-level validation evidence, with accuracy of 0.792–0.800 on scenes excluded from gradient updates. However, the five validation scenes informed checkpoint selection and therefore do not constitute an untouched test set. The primary evaluation against dense human A/B/Tie labels covers VIFB's 21 scenes and 25 methods. Section 4.8 extends the evaluation to EVAFusion through zero-shot transfer and fine-tuning experiments with a scene-disjoint test split and Overall-derived proxy labels. These experiments assess transfer and adaptation across datasets and annotation protocols; direct validation against dense human pairwise judgments beyond VIFB remains open. The third concern is that thousands of pairs might overstate the amount of independent evidence. Within the VIFB corpus, comparisons are clustered within 21 scenes, making the scene the appropriate sampling unit for uncertainty estimation. Scene-level confidence intervals remain necessary to quantify uncertainty while accounting for this dependence.

The consistency diagnostics of Section 4.7 additionally establish that LPIFM behaves like a measure in the ordinal sense: its verdicts are antisymmetric under candidate swap (100% swap agreement on U-I/All-M, ≥99.98% on all splits), free of decisive preference cycles (0 of 10,054 triplets, against a human rate of 0.0074), and fully transitive (hard and weak transitivity 1.0 versus human 0.9624/0.9786). We therefore describe LPIFM as a measure-like preference instrument. This internal coherence reflects the near-total order induced by the frozen decoder and is orthogonal to agreement with human labels (pairwise accuracy 0.792 on U-I/All-M); the corresponding left–right symmetry of the human panel is not identifiable from unordered gold labels.

Several limitations bound the claims. First, external validity: it is partially addressed by the cross-protocol EVAFusion study in Section 4.8, but that study relies on Overall-derived proxy labels, a single split seed, and one external corpus, so validation against dense human pairwise labels beyond VIFB remains to be demonstrated. Second, tie calibration: $F1_{Tie}$ (0.215–0.268 on All-M slices) shows ties remain the hardest class, and tie support on small held-out slices is too sparse to assess at all. Third, scene-level reliability: the Fig. S4 failure case (GTF on tricycle) shows that per-scene decisions can misfire when scene and method shift combine, so per-scene outputs should be treated with more caution than pool-level rankings. Fourth, most ablation variants are single runs, and the viewing-condition records of the subjective study were not retained. These limitations motivate further human-labeled external validation, scene-level bootstrap confidence intervals, and repeated-seed ablation studies.

LPIFM is a surrogate: it reproduces the recorded consensus of one carefully collected human panel, on one benchmark, for one fusion task. New domains, new artifact regimes, and high-stakes applications will still require human validation, and the surrogate should be re-anchored as preference data accumulate. The surrogate changes the cost structure of that loop: human effort can be reserved for anchoring and auditing, while routine method comparison, ablation triage, and leaderboard construction run at machine cost with human-aligned semantics.

## 6. CONCLUSION

LPIFM formulates infrared–visible fusion assessment as source-conditioned A/B/Tie prediction from both source images and both fused candidates. Its public preference corpus contains all 6,300 adjudicated unordered comparisons among 25 methods on 21 VIFB scenes. Across four evaluation settings, LPIFM achieves 79.2–84.0% pairwise accuracy and ranking correlations of 0.941–0.977. On the two full method pools, it outperforms the strongest of 19 conventional metrics by 16.3–21.1 pp in accuracy. On U-I/All-M, it reduces mean absolute rank difference to 1.68, compared with 5.20 for the strongest conventional baseline. Strong ordinal consistency on the evaluated VIFB splits further supports stable pair decisions and method rankings under the evaluated protocol.

On the scene-disjoint EVAFusion proxy test, three-epoch adaptation reaches 0.5066 accuracy, 0.4941 macro-$F1$, and mean within-scene $\rho = 0.4449$, outperforming all 19 conventional metrics on all three primary measures. This result demonstrates rapid adaptation to the distinct score-derived preference convention examined here. Direct validation against dense human pairwise judgments beyond VIFB remains future work. LPIFM thereby provides a practical route from direct human comparisons to scalable fusion-method evaluation. Future work should collect dense human preferences in other fusion domains and explore preference-driven optimization of fusion models. The released preference corpus, model weights, and code provide a public basis for auditing rankings, retraining the comparator, and extending human-aligned fusion assessment.

## ETHICS STATEMENT

The subjective annotation study involved adult volunteers performing non-invasive image preference judgments and did not collect personally identifiable information. Under the institutional guidelines of Chengdu University of Technology, ethical approval was not required for this type of minimal-risk preference annotation research. All participants were informed of the study purpose and procedure and provided informed consent before taking part. Participation was voluntary, with the right to withdraw at any time, and no personally identifiable information was collected or disclosed.

## FUNDING

This research did not receive any specific grant from funding agencies in the public, commercial, or not-for-profit sectors.

## AUTHOR CONTRIBUTIONS

Haoran Liu: Conceptualization, Formal analysis, Investigation, Methodology, Visualization, Writing – original draft, Writing – review and editing. Mingzhe Liu: Data curation, Project administration, Resources, Supervision. Peng Li: Software, Validation, Visualization. Guibin Zan: Project administration, Resources, Supervision.

## DECLARATION OF GENERATIVE AI AND AI-ASSISTED TECHNOLOGIES IN THE MANUSCRIPT PREPARATION PROCESS

During the preparation of this work, the authors used Claude Fable 5 and GPT-6 Astra to improve the language and readability of the manuscript. After using these tools, the authors reviewed and edited the content as needed and take full responsibility for the content of the published article.

## DATA AVAILABILITY

The annotated pairwise preference dataset (6,300 adjudicated unordered comparisons, 13,125 ordered corpus records, and scene/method split manifests), together with the LPIFM model weights, source code, and evaluation code, is made publicly available at https://github.com/HaoranLiu507/LPIFM. The source code is released under the GNU Affero General Public License v3.0 (AGPL-3.0), and the model weights and datasets are released under the Creative Commons Attribution Non Commercial Share Alike 4.0 International (CC BY-NC-SA 4.0) license.

# Supplementary Material

## Ranking Image Fusion the Way Humans Do: A Learned Pairwise Preference Measure for Infrared–Visible Fusion Assessment

Haoran Liu[1,2], Mingzhe Liu[1,2,*], Peng Li[2], Guibin Zan[3,*]

1 School of Artificial Intelligence and Electronic Engineering, Sichuan Technology and Business University, Chengdu 611745, China

2 College of Nuclear Technology and Automation Engineering, Chengdu University of Technology, Chengdu 610059, China

3 National Synchrotron Radiation Laboratory, University of Science and Technology of China, Hefei 230029, China

*Corresponding author, liumz@cdut.edu.cn (Mingzhe Liu), zangb@mail.ustc.edu.cn (Guibin Zan)

This document provides supplementary figures (Figs. S1–S7), tables (Tables S1–S9), and additional analyses (Sections S1 and S2) supporting the main text. Setting abbreviations follow the main text: U = unseen, All = all; I = images, M = methods. All objective-metric decisions use the frozen tie-band calibration $\kappa^* = 0.09$ selected on the training seen-scene × seen-method split; LPIFM uses its own fixed threshold $t = 0.3$.

## Supplementary Figures

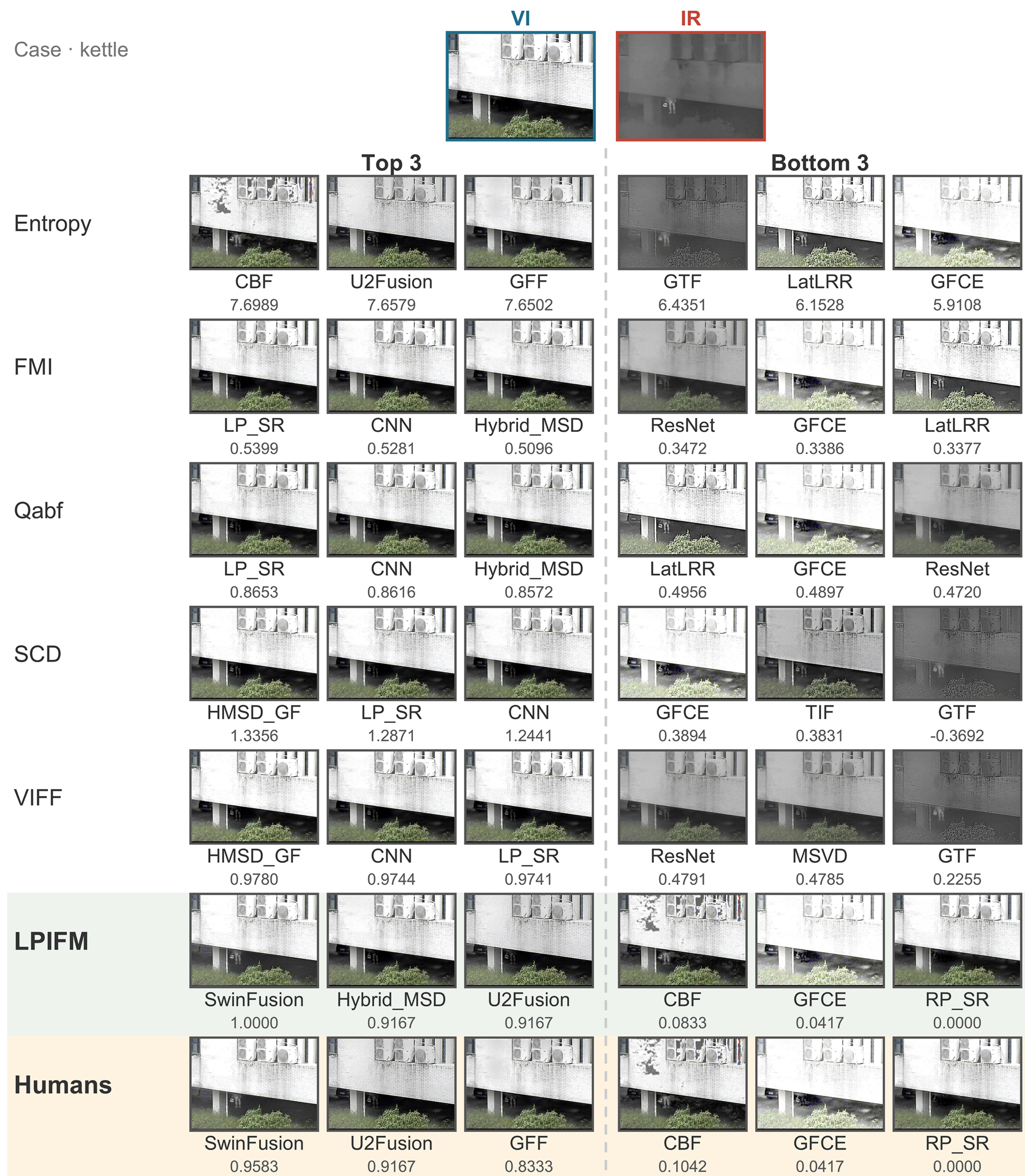

Fig. S1. Per-scene ranking case study on the kettle scene: for each evaluator (rows: Entropy, FMI, Qabf, SCD, VIFF, LPIFM, Humans), the three highest-ranked and three lowest-ranked fused results are shown with the corresponding scores (classical metrics: native scalar values; LPIFM and Humans: per-scene normalized win rates). LPIFM and human observers agree on SwinFusion as the scene champion.

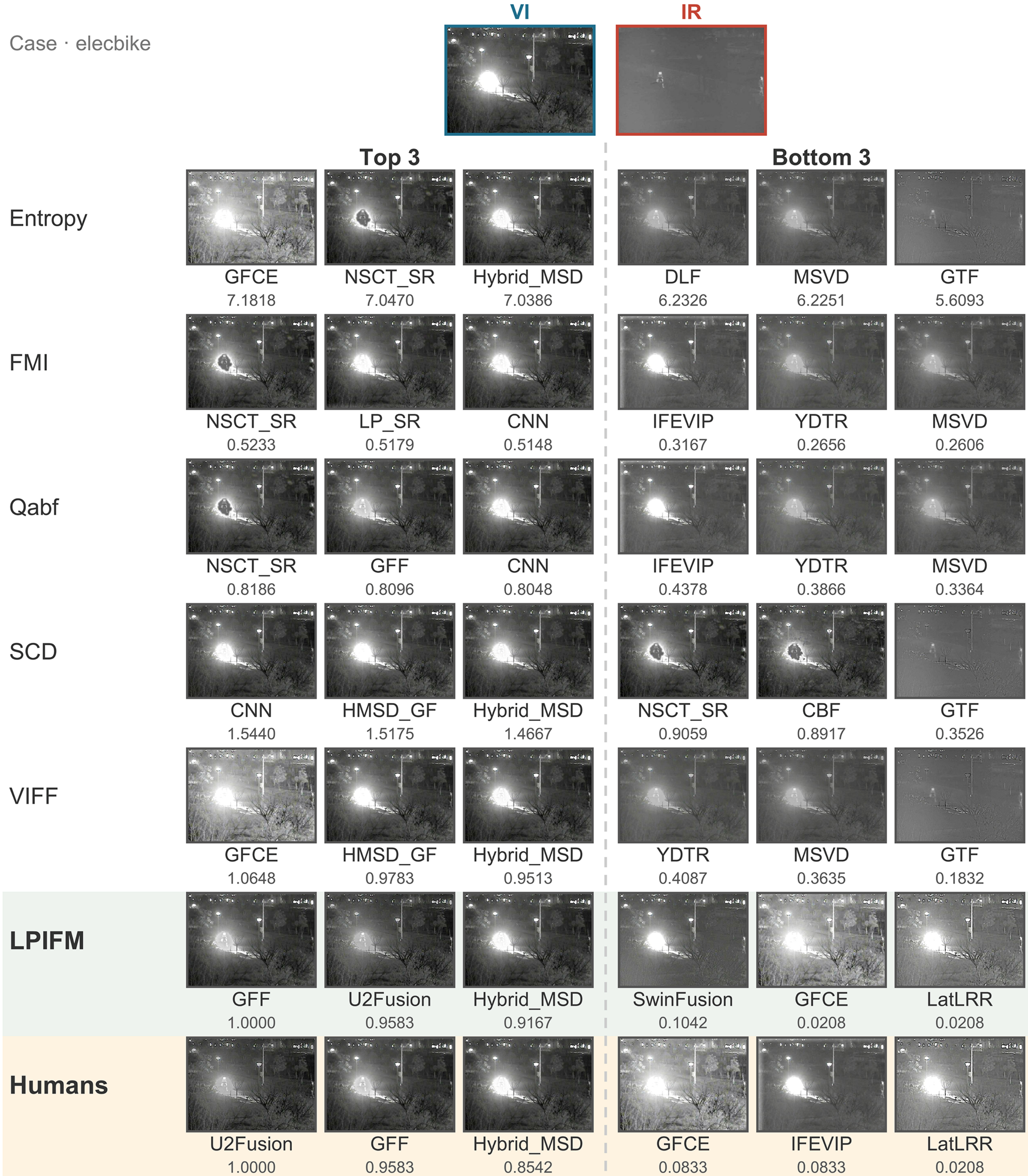

Fig. S2. Per-scene ranking case study on the elecbike scene, in the format of Fig. S1. The LPIFM and human top-three sets are identical.

Case · labMan

VI IR

| | Top 3 | | | Bottom 3 | | |
|---|---|---|---|---|---|---|
| Entropy | HMSD_GF 7.7410 | GFCE 7.7219 | IFEVIP 7.7124 | YDTR 6.7956 | LatLRR 6.7133 | GTF 5.6731 |
| FMI | LatLRR 0.3585 | VSMWLS 0.3262 | IFCNN 0.3188 | FPDE 0.2638 | ADF 0.2629 | MSVD 0.2565 |
| Qabf | NSCT_SR 0.6156 | GFF 0.6125 | HMSD_GF 0.6066 | ResNet 0.3876 | YDTR 0.3611 | MSVD 0.2335 |
| SCD | VSMWLS 1.6782 | SwinFusion 1.6760 | CNN 1.6249 | GFF 0.9811 | NSCT_SR 0.6244 | CBF 0.6136 |
| VIFF | LatLRR 0.8231 | HMSD_GF 0.7528 | GFCE 0.7208 | FPDE 0.4237 | MSVD 0.3675 | GTF 0.3227 |
| **LPIFM** | IFEVIP 1.0000 | SwinFusion 0.9583 | SeAFusion 0.9167 | GFCE 0.0833 | CBF 0.0417 | RP_SR 0.0000 |
| **Humans** | SeAFusion 0.9375 | SwinFusion 0.9375 | ADF 0.8750 | NSCT_SR 0.0833 | RP_SR 0.0833 | GTF 0.0000 |

Fig. S3. Per-scene ranking case study on the labMan scene, in the format of Fig. S1.

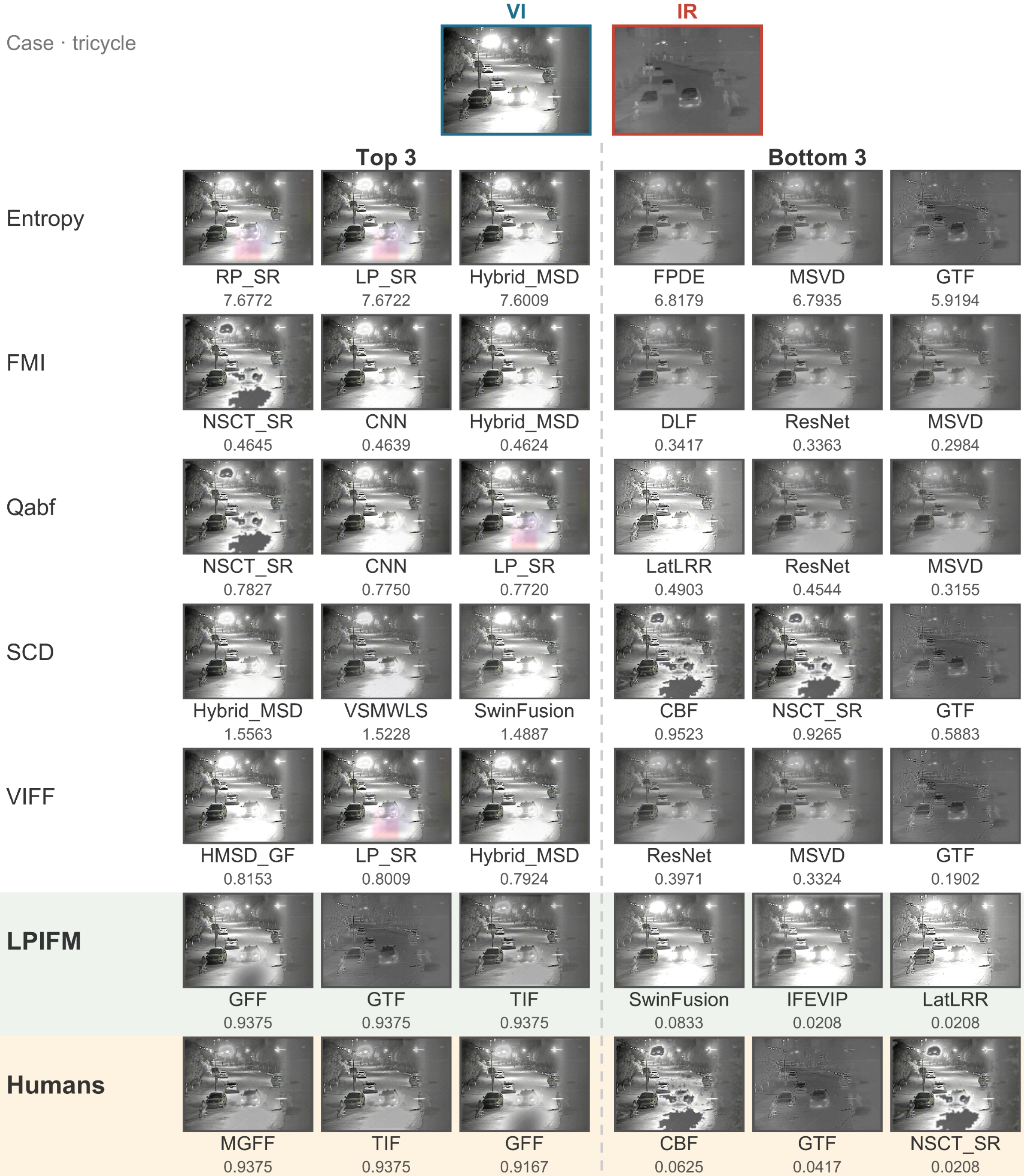

Fig. S4. Per-scene ranking case study on the tricycle scene, in the format of Fig. S1 (failure case). LPIFM places GTF in its leading group (0.9375) while human observers rank GTF near the bottom (0.0417). The tricycle scene is a validation scene and GTF is a held-out method, so this case combines both generalization factors; its implication for scene-level reliability is discussed in Section 5 of the main text.

Case · walking

VI IR

| | Top 3 | | | Bottom 3 | | |
|---|---|---|---|---|---|---|
| Entropy | CNN 7.7844 | HMSD_GF 7.7604 | NSCT_SR 7.7322 | FPDE 7.1287 | DLF 7.1156 | MSVD 7.0687 |
| FMI | GFF 0.5193 | NSCT_SR 0.4611 | ADF 0.4600 | TIF 0.3645 | CBF 0.3601 | MSVD 0.2809 |
| Qabf | GFF 0.6047 | NSCT_SR 0.5571 | CNN 0.4980 | DLF 0.3654 | ResNet 0.3495 | MSVD 0.2751 |
| SCD | HMSD_GF 1.7676 | IFCNN 1.7607 | CNN 1.7475 | CBF 1.1143 | NSCT_SR 1.0826 | GTF 0.9069 |
| VIFF | MGFF 0.4488 | LatLRR 0.3977 | U2Fusion 0.3875 | CBF 0.1721 | GFF 0.1682 | GTF 0.1641 |
| **LPIFM** | IFCNN 1.0000 | SeAFusion 0.9375 | U2Fusion 0.9375 | MSVD 0.0833 | NSCT_SR 0.0417 | CBF 0.0000 |
| **Humans** | U2Fusion 0.9583 | IFCNN 0.9375 | SwinFusion 0.9167 | MSVD 0.1042 | CBF 0.0417 | GTF 0.0000 |

Fig. S5. Per-scene ranking case study on the walking scene, in the format of Fig. S1. LPIFM's top and bottom sets each share two of three members with the human sets, while classical metrics diverge far more strongly.

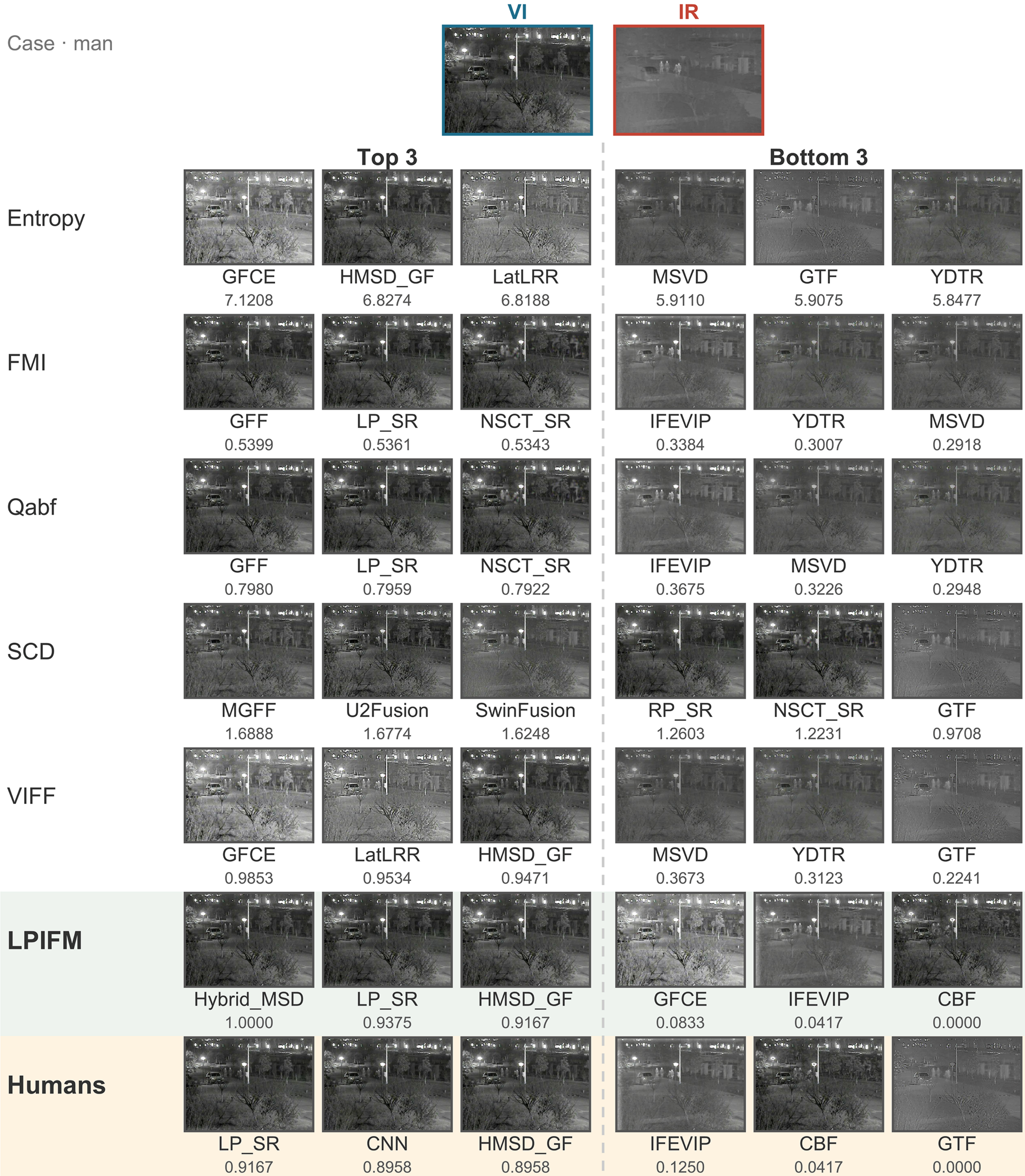

Fig. S6. Per-scene ranking case study on the man scene, in the format of Fig. S1. LPIFM and human observers agree on LP_SR and HMSD_GF as the leading group in this night scene, while several classical metrics rank strongly artifacted results highly.

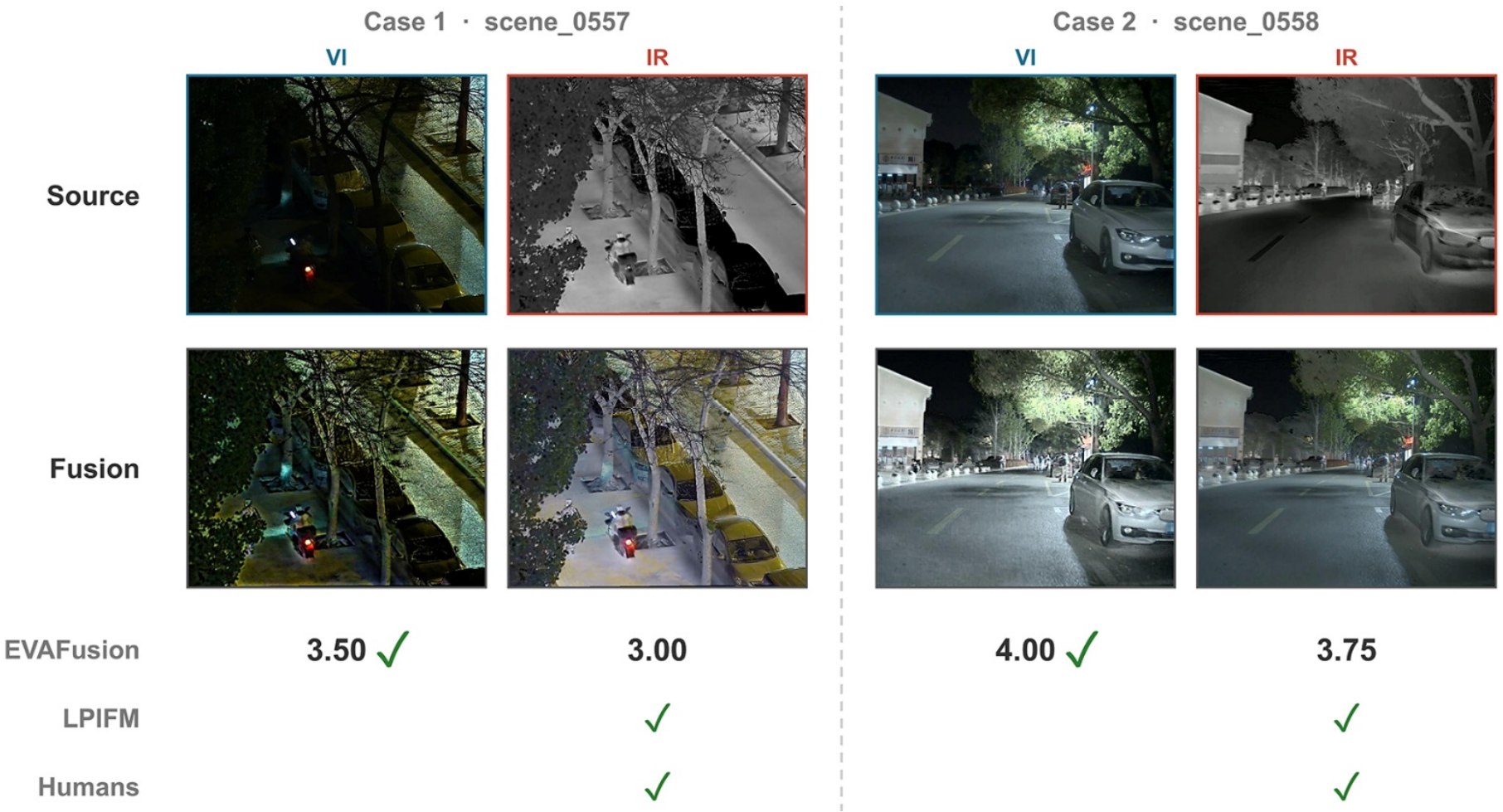

Fig. S7. Disagreement between EVAFusion Overall-derived labels and source-conditioned preferences on two held-out pairs (scene_0557 and scene_0558). The higher Overall score defines the proxy label; check marks in the LPIFM and Humans rows indicate their preferred candidates. Zero-shot LPIFM agrees with humans in both cases.

## Supplementary Tables

Table S1. Full pairwise classification and T-BT rank correlation on unseen images × unseen methods (U-I/U-M).

| Method | Acc↑ | $F1_A$↑ | $F1_B$↑ | $\rho$(T-BT)↑ | $\tau$(T-BT)↑ |
|---|---|---|---|---|---|
| Avg_gradient | 0.547 | 0.108 | 0.712 | 0.143 | 0.333 |
| CC | 0.680 | 0.698 | 0.709 | 0.600 | 0.467 |
| Cross_entropy | 0.440 | 0.255 | 0.571 | -0.086 | -0.067 |
| Edge_intensity | 0.547 | 0.108 | 0.712 | 0.143 | 0.333 |
| Entropy | 0.533 | 0.261 | 0.673 | 0.086 | 0.200 |
| FMI | 0.453 | 0.441 | 0.514 | 0.086 | 0.067 |
| MS-SSIM | 0.640 | 0.740 | 0.609 | 0.600 | 0.467 |
| Mutinf | 0.400 | 0.361 | 0.456 | -0.371 | -0.333 |
| Nabf | 0.280 | 0.430 | 0.040 | -0.609 | -0.414 |
| Psnr | 0.600 | 0.707 | 0.508 | 0.257 | 0.200 |
| Qabf | 0.547 | 0.448 | 0.644 | -0.029 | 0.067 |
| Qcb | 0.613 | 0.526 | 0.705 | 0.200 | 0.200 |
| Qcv | 0.720 | 0.774 | 0.759 | 0.771 | 0.600 |
| Rmse | 0.600 | 0.707 | 0.508 | 0.257 | 0.200 |
| SCD | 0.840 | 0.778 | 0.891 | 0.943 | 0.867 |
| Spatial_frequency | 0.533 | 0.190 | 0.713 | 0.029 | 0.067 |
| Ssim | 0.613 | 0.683 | 0.581 | 0.200 | 0.067 |
| Variance | 0.627 | 0.458 | 0.737 | 0.543 | 0.333 |
| VIFF | 0.653 | 0.531 | 0.774 | 0.725 | 0.552 |
| LPIFM | 0.800 | 0.772 | 0.844 | 0.943 | 0.867 |

*Note: frozen $\kappa^* = 0.09$ (calibrated on the training seen×seen split; $r_{tie}(\kappa^*)$ = 0.0624 vs. ground-truth 0.0614, relative error 0.0163). Acc is three-class accuracy including ties. Ground-truth ties in this setting: 2/75, below the reporting threshold of 30, so $F1_{Tie}$ and macro-$F1$ are omitted (see Section 4.1 of the main text).*

Table S2. Full pairwise classification and T-BT rank correlation on all images × unseen methods (All-I/U-M).

| Method | Acc↑ | $F1_A$↑ | $F1_B$↑ | $\rho$(T-BT)↑ | $\tau$(T-BT)↑ |
|---|---|---|---|---|---|
| Avg_gradient | 0.590 | 0.263 | 0.753 | 0.429 | 0.467 |
| CC | 0.676 | 0.704 | 0.709 | 0.600 | 0.467 |
| Cross_entropy | 0.438 | 0.284 | 0.556 | -0.143 | -0.067 |
| Edge_intensity | 0.597 | 0.266 | 0.755 | 0.429 | 0.467 |
| Entropy | 0.594 | 0.390 | 0.721 | 0.319 | 0.276 |
| FMI | 0.441 | 0.413 | 0.503 | -0.257 | -0.200 |
| MS-SSIM | 0.616 | 0.660 | 0.623 | 0.429 | 0.200 |
| Mutinf | 0.429 | 0.378 | 0.506 | -0.257 | -0.200 |
| Nabf | 0.260 | 0.392 | 0.078 | -0.771 | -0.600 |
| Psnr | 0.584 | 0.644 | 0.565 | 0.371 | 0.067 |
| Qabf | 0.584 | 0.425 | 0.712 | 0.257 | 0.200 |
| Qcb | 0.619 | 0.474 | 0.738 | 0.429 | 0.333 |
| Qcv | 0.686 | 0.696 | 0.753 | 0.829 | 0.733 |
| Rmse | 0.584 | 0.642 | 0.567 | 0.371 | 0.067 |
| SCD | 0.803 | 0.785 | 0.851 | 0.943 | 0.867 |
| Spatial_frequency | 0.581 | 0.319 | 0.733 | 0.429 | 0.333 |
| Ssim | 0.556 | 0.626 | 0.535 | 0.086 | -0.067 |
| Variance | 0.603 | 0.414 | 0.734 | 0.429 | 0.333 |
| VIFF | 0.708 | 0.609 | 0.805 | 0.771 | 0.600 |
| LPIFM | 0.813 | 0.800 | 0.865 | 0.943 | 0.867 |

*Note: same protocol as Table S1. Ground-truth ties in this setting: 13/315, below the reporting threshold of 30, so $F1_{Tie}$ and macro-$F1$ are omitted.*

Table S3. T-BT rank table on all images × all methods (All-I/All-M): fusion-method rankings induced by each evaluator (rank 1 is best).

| Rank | Humans | LPIFM | SCD | FMI | Qcv | Psnr | Qcb | Qabf |
|---|---|---|---|---|---|---|---|---|
| 1 | Hybrid_MSD | U2Fusion | U2Fusion | LP_SR | SwinFusion | DLF | LP_SR | LP_SR |
| 2 | U2Fusion | Hybrid_MSD | IFCNN | CNN | CNN | ResNet | Hybrid_MSD | CNN |
| 3 | IFCNN | IFCNN | HMSD_GF | GFF | SeAFusion | ADF | GFF | NSCT_SR |
| 4 | MGFF | TIF | MGFF | NSCT_SR | Hybrid_MSD | MSVD | CNN | Hybrid_MSD |
| 5 | TIF | GFF | VSMWLS | Hybrid_MSD | IFCNN | FPDE | HMSD_GF | GFF |
| 6 | HMSD_GF | MGFF | SwinFusion | ADF | HMSD_GF | IFCNN | NSCT_SR | HMSD_GF |
| 7 | GFF | HMSD_GF | LatLRR | IFCNN | LP_SR | TIF | RP_SR | IFCNN |
| 8 | CNN | CNN | CNN | HMSD_GF | TIF | MGFF | MGFF | TIF |
| 9 | VSMWLS | VSMWLS | Hybrid_MSD | FPDE | IFEVIP | VSMWLS | TIF | SwinFusion |
| 10 | LP_SR | LP_SR | YDTR | MGFF | YDTR | Hybrid_MSD | U2Fusion | CBF |
| 11 | SwinFusion | ADF | SeAFusion | SwinFusion | VSMWLS | GFF | GFCE | SeAFusion |
| 12 | ADF | SwinFusion | TIF | VSMWLS | GFF | YDTR | IFCNN | MGFF |
| 13 | SeAFusion | DLF | DLF | IFEVIP | MGFF | U2Fusion | CBF | RP_SR |
| 14 | DLF | ResNet | GFCE | GTF | ResNet | LP_SR | VSMWLS | VSMWLS |
| 15 | ResNet | FPDE | ResNet | RP_SR | LatLRR | CNN | LatLRR | U2Fusion |
| 16 | FPDE | SeAFusion | FPDE | LatLRR | U2Fusion | HMSD_GF | SwinFusion | ADF |
| 17 | YDTR | YDTR | ADF | GFCE | DLF | GTF | ADF | IFEVIP |
| 18 | MSVD | RP_SR | LP_SR | SeAFusion | RP_SR | RP_SR | FPDE | GFCE |
| 19 | GFCE | MSVD | MSVD | U2Fusion | ADF | CBF | SeAFusion | FPDE |
| 20 | RP_SR | GTF | IFEVIP | TIF | FPDE | NSCT_SR | IFEVIP | YDTR |
| 21 | LatLRR | GFCE | GFF | YDTR | MSVD | SeAFusion | ResNet | GTF |
| 22 | IFEVIP | LatLRR | RP_SR | DLF | NSCT_SR | SwinFusion | DLF | LatLRR |
| 23 | NSCT_SR | NSCT_SR | CBF | CBF | GFCE | IFEVIP | YDTR | DLF |
| 24 | CBF | IFEVIP | NSCT_SR | ResNet | CBF | GFCE | GTF | ResNet |
| 25 | GTF | CBF | GTF | MSVD | GTF | LatLRR | MSVD | MSVD |

*Note: ranks follow descending T-BT strength; ties are broken alphabetically. Humans denotes the ranking derived from ground-truth labels. LPIFM–Humans MARD = 1.20; top-1 agreement: no (adjacent-leader swap); top-3 overlap: 3/3. The best objective MARD in this table is SCD at 3.68; the worst is FMI at 6.72.*

Table S4. Robustness of the ranking conclusions to the aggregation rule: Spearman $\boldsymbol{\rho}$ and Kendall $\tau_b$ between predicted and human method vectors under top-share (Ti), normalized win rate (T-NR), and normalized win share (T-NW) aggregation, for all four evaluation settings.

**(a) U-I/U-M**

| Method | $\rho$(Ti)↑ | $\tau$(Ti)↑ | $\rho$(T-NR)↑ | $\tau$(T-NR)↑ | $\rho$(T-NW)↑ | $\tau$(T-NW)↑ |
|---|---|---|---|---|---|---|
| Avg_gradient | 0.072 | 0.096 | 0.143 | 0.333 | 0.143 | 0.333 |
| CC | 0.098 | 0.087 | 0.600 | 0.467 | 0.600 | 0.467 |
| Cross_entropy | -0.417 | -0.387 | -0.086 | -0.067 | -0.086 | -0.067 |
| Edge_intensity | 0.072 | 0.096 | 0.143 | 0.333 | 0.143 | 0.333 |
| Entropy | 0.721 | 0.609 | 0.086 | 0.200 | 0.086 | 0.200 |
| FMI | -0.197 | -0.174 | 0.086 | 0.067 | 0.086 | 0.067 |
| MS-SSIM | 0.072 | 0.096 | 0.551 | 0.414 | 0.551 | 0.414 |
| Mutinf | -0.484 | -0.417 | -0.371 | -0.333 | -0.371 | -0.333 |
| Nabf | -0.647 | -0.577 | -0.609 | -0.414 | -0.609 | -0.414 |
| Psnr | -0.417 | -0.387 | 0.257 | 0.200 | 0.257 | 0.200 |
| Qabf | 0.000 | 0.000 | 0.087 | 0.138 | 0.087 | 0.138 |
| Qcb | 0.108 | 0.096 | 0.200 | 0.200 | 0.200 | 0.200 |
| Qcv | 0.844 | 0.721 | 0.771 | 0.600 | 0.771 | 0.600 |
| Rmse | -0.417 | -0.387 | 0.257 | 0.200 | 0.257 | 0.200 |
| SCD | 0.898 | 0.866 | 0.943 | 0.867 | 0.943 | 0.867 |
| Spatial_frequency | 0.426 | 0.435 | 0.029 | 0.067 | 0.029 | 0.067 |
| Ssim | -0.066 | -0.087 | 0.200 | 0.067 | 0.200 | 0.067 |
| Variance | 0.557 | 0.435 | 0.543 | 0.333 | 0.543 | 0.333 |
| VIFF | 0.893 | 0.772 | 0.725 | 0.552 | 0.725 | 0.552 |
| LPIFM | 0.742 | 0.667 | 0.943 | 0.867 | 0.943 | 0.867 |

**(b) U-I/All-M**

| Method | $\rho$(Ti)↑ | $\tau$(Ti)↑ | $\rho$(T-NR)↑ | $\tau$(T-NR)↑ | $\rho$(T-NW)↑ | $\tau$(T-NW)↑ |
|---|---|---|---|---|---|---|
| Avg_gradient | -0.226 | -0.212 | -0.015 | 0.007 | -0.015 | 0.007 |
| CC | 0.007 | 0.007 | 0.412 | 0.312 | 0.412 | 0.312 |
| Cross_entropy | 0.147 | 0.145 | 0.121 | 0.097 | 0.121 | 0.097 |
| Edge_intensity | -0.226 | -0.212 | 0.018 | 0.034 | 0.018 | 0.034 |
| Entropy | 0.225 | 0.209 | 0.192 | 0.121 | 0.192 | 0.121 |
| FMI | 0.273 | 0.253 | 0.345 | 0.224 | 0.345 | 0.224 |
| MS-SSIM | -0.304 | -0.278 | 0.431 | 0.309 | 0.431 | 0.309 |
| Mutinf | 0.008 | 0.009 | 0.058 | 0.071 | 0.058 | 0.071 |
| Nabf | -0.376 | -0.338 | 0.099 | 0.077 | 0.099 | 0.077 |
| Psnr | -0.307 | -0.292 | 0.549 | 0.335 | 0.549 | 0.335 |
| Qabf | 0.163 | 0.151 | 0.426 | 0.274 | 0.426 | 0.274 |
| Qcb | 0.375 | 0.360 | 0.503 | 0.355 | 0.503 | 0.355 |
| Qcv | 0.464 | 0.392 | 0.473 | 0.345 | 0.473 | 0.345 |
| Rmse | -0.307 | -0.292 | 0.532 | 0.325 | 0.532 | 0.325 |
| SCD | 0.588 | 0.566 | 0.559 | 0.419 | 0.559 | 0.419 |
| Spatial_frequency | -0.181 | -0.170 | -0.043 | -0.037 | -0.043 | -0.037 |
| Ssim | -0.181 | -0.172 | 0.442 | 0.295 | 0.442 | 0.295 |
| Variance | 0.722 | 0.648 | 0.088 | 0.067 | 0.088 | 0.067 |
| VIFF | 0.440 | 0.387 | 0.381 | 0.251 | 0.381 | 0.251 |
| LPIFM | 0.453 | 0.428 | 0.940 | 0.826 | 0.940 | 0.826 |

**(c) All-I/U-M**

| Method | $\rho$(Ti)↑ | $\tau$(Ti)↑ | $\rho$(T-NR)↑ | $\tau$(T-NR)↑ | $\rho$(T-NW)↑ | $\tau$(T-NW)↑ |
|---|---|---|---|---|---|---|
| Avg_gradient | 0.250 | 0.357 | 0.429 | 0.467 | 0.429 | 0.467 |
| CC | 0.524 | 0.463 | 0.600 | 0.467 | 0.600 | 0.467 |
| Cross_entropy | -0.582 | -0.445 | -0.143 | -0.067 | -0.143 | -0.067 |
| Edge_intensity | 0.388 | 0.445 | 0.429 | 0.467 | 0.429 | 0.467 |
| Entropy | 0.029 | 0.000 | 0.319 | 0.276 | 0.319 | 0.276 |
| FMI | -0.754 | -0.552 | -0.257 | -0.200 | -0.257 | -0.200 |
| MS-SSIM | 0.824 | 0.714 | 0.429 | 0.200 | 0.429 | 0.200 |
| Mutinf | -0.716 | -0.593 | -0.257 | -0.200 | -0.257 | -0.200 |
| Nabf | -0.223 | -0.178 | -0.771 | -0.600 | -0.771 | -0.600 |
| Psnr | 0.377 | 0.267 | 0.371 | 0.067 | 0.371 | 0.067 |
| Qabf | -0.223 | -0.178 | 0.257 | 0.200 | 0.257 | 0.200 |
| Qcb | 0.309 | 0.214 | 0.429 | 0.333 | 0.429 | 0.333 |
| Qcv | -0.116 | 0.000 | 0.829 | 0.733 | 0.829 | 0.733 |
| Rmse | 0.377 | 0.267 | 0.371 | 0.067 | 0.371 | 0.067 |
| SCD | 0.832 | 0.772 | 0.943 | 0.867 | 0.943 | 0.867 |
| Spatial_frequency | 0.191 | 0.214 | 0.429 | 0.333 | 0.429 | 0.333 |
| Ssim | 0.074 | 0.071 | 0.086 | -0.067 | 0.086 | -0.067 |
| Variance | 0.232 | 0.138 | 0.429 | 0.333 | 0.429 | 0.333 |

| Method | $\rho$(Ti)↑ | $\tau$(Ti)↑ | $\rho$(T-NR)↑ | $\tau$(T-NR)↑ | $\rho$(T-NW)↑ | $\tau$(T-NW)↑ |
|---|---|---|---|---|---|---|
| VIFF | 0.426 | 0.357 | 0.771 | 0.600 | 0.771 | 0.600 |
| LPIFM | 0.868 | 0.786 | 0.943 | 0.867 | 0.943 | 0.867 |

**(d) All-I/All-M**

| Method | $\rho$(Ti)↑ | $\tau$(Ti)↑ | $\rho$(T-NR)↑ | $\tau$(T-NR)↑ | $\rho$(T-NW)↑ | $\tau$(T-NW)↑ |
|---|---|---|---|---|---|---|
| Avg_gradient | -0.196 | -0.171 | 0.072 | 0.107 | 0.072 | 0.107 |
| CC | -0.140 | -0.103 | 0.385 | 0.273 | 0.385 | 0.273 |
| Cross_entropy | -0.148 | -0.131 | 0.244 | 0.197 | 0.244 | 0.197 |
| Edge_intensity | -0.196 | -0.171 | 0.081 | 0.127 | 0.081 | 0.127 |
| Entropy | 0.323 | 0.258 | 0.230 | 0.187 | 0.230 | 0.187 |
| FMI | 0.077 | 0.059 | 0.362 | 0.230 | 0.362 | 0.230 |
| MS-SSIM | 0.156 | 0.148 | 0.375 | 0.207 | 0.375 | 0.207 |
| Mutinf | 0.022 | 0.017 | 0.168 | 0.140 | 0.168 | 0.140 |
| Nabf | -0.388 | -0.330 | -0.056 | -0.060 | -0.056 | -0.060 |
| Psnr | -0.446 | -0.389 | 0.422 | 0.260 | 0.422 | 0.260 |
| Qabf | 0.267 | 0.221 | 0.479 | 0.313 | 0.479 | 0.313 |
| Qcb | 0.472 | 0.432 | 0.515 | 0.364 | 0.515 | 0.364 |
| Qcv | 0.514 | 0.438 | 0.642 | 0.460 | 0.642 | 0.460 |
| Rmse | -0.499 | -0.431 | 0.417 | 0.250 | 0.417 | 0.250 |
| SCD | 0.604 | 0.516 | 0.717 | 0.560 | 0.717 | 0.560 |
| Spatial_frequency | -0.004 | -0.008 | 0.080 | 0.120 | 0.080 | 0.120 |
| Ssim | -0.446 | -0.389 | 0.312 | 0.187 | 0.312 | 0.187 |
| Variance | 0.405 | 0.345 | 0.173 | 0.147 | 0.173 | 0.147 |
| VIFF | 0.556 | 0.448 | 0.502 | 0.387 | 0.502 | 0.387 |
| LPIFM | 0.526 | 0.415 | 0.976 | 0.898 | 0.976 | 0.898 |

*Note: shared frozen $\kappa^* = 0.09$ with the main tables. Objective-side Ti/T-* vectors are aggregated from the $\kappa \cdot MAD$ ternary decisions, not from raw scalar scores. LPIFM's $\rho$(T-NR)/$\rho$(T-NW) remain 0.940–0.976 across settings, so the main-text ranking conclusions are a property of the predicted decisions, not of one aggregation rule.*

Table S5. $\kappa$ calibration grid on the training seen-scene × seen-method split, and per-setting ground-truth tie rates.

**(a) Calibration grid (calibration-set ground-truth tie rate 0.0614).**

| $\kappa$ | $r_{tie}(\kappa)$ | $\lvert r_{tie}(\kappa) - r_{tie\ GT}\rvert$ | **Selected** |
|---|---|---|---|
| 0.01 | 0.0078 | 0.0536 | |
| 0.02 | 0.0153 | 0.0461 | |
| 0.03 | 0.0221 | 0.0393 | |
| 0.04 | 0.0292 | 0.0322 | |
| 0.05 | 0.0362 | 0.0252 | |
| 0.06 | 0.0432 | 0.0182 | |
| 0.07 | 0.0499 | 0.0115 | |
| 0.08 | 0.0561 | 0.0053 | |
| **0.09** | **0.0624** | **0.0010** | ✓ ($\kappa^*$) |
| 0.10 | 0.0684 | 0.0070 | |
| 0.15 | 0.0994 | 0.0380 | |
| 0.20 | 0.1300 | 0.0686 | |
| 0.30 | 0.1862 | 0.1248 | |

**(b) Ground-truth tie rates per evaluation setting (descriptive only; the information is never used to select $\kappa^*$).**

| **Setting** | **Frozen $\kappa^*$** | $r_{tie\ GT}$ **(evaluation)** | $r_{tie\ GT}$ **(calibration)** |
|---|---|---|---|
| U-I/U-M | 0.09 | 0.0267 | 0.0614 |
| U-I/All-M | 0.09 | 0.0620 | 0.0614 |
| All-I/U-M | 0.09 | 0.0413 | 0.0614 |
| All-I/All-M | 0.09 | 0.0656 | 0.0614 |

*Note: $\kappa^* = 0.09$ was selected once on the training seen×seen calibration split (19 methods × 16 scenes) by matching the objective tie rate $r_{tie}(\kappa)$ to the calibration-set ground-truth tie rate (relative error 0.0163 ≤ 10%), then frozen across all four reported settings. Re-selecting $\kappa$ per test setting is a test-informed oracle and is excluded from the main results.*

Table S6. Pairwise classification and T-BT rank correlation on all images × all methods (All-I/All-M).

| **Method** | **Acc↑** | $F1_A$↑ | $F1_B$↑ | $F1_{Tie}$↑ | **macro-**$F1$↑ | $\rho$**(T-BT)↑** | $\tau$**(T-BT)↑** |
|---|---|---|---|---|---|---|---|
| Avg_gradient | 0.478 | 0.478 | 0.517 | 0.180 | 0.392 | 0.072 | 0.107 |
| CC | 0.543 | 0.515 | 0.598 | 0.294 | 0.469 | 0.385 | 0.273 |
| Cross_entropy | 0.453 | 0.466 | 0.473 | 0.196 | 0.378 | 0.244 | 0.197 |
| Edge_intensity | 0.479 | 0.478 | 0.519 | 0.185 | 0.394 | 0.081 | 0.127 |
| Entropy | 0.533 | 0.528 | 0.566 | 0.328 | 0.474 | 0.237 | 0.193 |
| FMI | 0.540 | 0.566 | 0.567 | 0.118 | 0.417 | 0.362 | 0.230 |
| MS-SSIM | 0.570 | 0.540 | 0.626 | 0.306 | 0.491 | 0.375 | 0.207 |
| Mutinf | 0.535 | 0.537 | 0.573 | 0.212 | 0.441 | 0.168 | 0.140 |
| Nabf | 0.455 | 0.473 | 0.468 | 0.216 | 0.386 | -0.056 | -0.060 |
| Psnr | 0.566 | 0.568 | 0.584 | 0.432 | 0.528 | 0.422 | 0.260 |
| Qabf | 0.576 | 0.587 | 0.618 | 0.153 | 0.453 | 0.479 | 0.313 |
| Qcb | 0.552 | 0.563 | 0.589 | 0.175 | 0.442 | 0.509 | 0.360 |
| Qcv | 0.605 | 0.589 | 0.666 | 0.253 | 0.502 | 0.642 | 0.460 |
| Rmse | 0.566 | 0.568 | 0.584 | 0.435 | 0.529 | 0.417 | 0.250 |
| SCD | 0.629 | 0.591 | 0.693 | 0.325 | 0.536 | 0.717 | 0.560 |
| Spatial_frequency | 0.473 | 0.479 | 0.512 | 0.142 | 0.377 | 0.081 | 0.124 |
| Ssim | 0.570 | 0.558 | 0.626 | 0.220 | 0.468 | 0.312 | 0.187 |
| Variance | 0.517 | 0.492 | 0.557 | 0.386 | 0.478 | 0.173 | 0.147 |
| VIFF | 0.587 | 0.575 | 0.636 | 0.265 | 0.492 | 0.502 | 0.387 |
| LPIFM | 0.840 | 0.859 | 0.876 | 0.268 | 0.667 | 0.977 | 0.900 |

*Note: same protocol as Table 2 of the main text. Ground-truth ties in this setting: 413/6,300.*

Table S7. T-BT rank table on unseen images × all methods (U-I/All-M): fusion-method rankings induced by each evaluator (rank 1 is best).

| Rank | Humans | LPIFM | SCD | FMI | Qcv | Psnr | Qcb | Qabf |
|---|---|---|---|---|---|---|---|---|
| 1 | U2Fusion | U2Fusion | U2Fusion | LP_SR | CNN | ADF | LP_SR | LP_SR |
| 2 | MGFF | Hybrid_MSD | HMSD_GF | CNN | HMSD_GF | DLF | Hybrid_MSD | CNN |
| 3 | TIF | TIF | IFCNN | Hybrid_MSD | IFCNN | FPDE | GFF | NSCT_SR |
| 4 | Hybrid_MSD | IFCNN | LatLRR | ADF | SwinFusion | MSVD | CNN | Hybrid_MSD |
| 5 | GFF | GFF | VSMWLS | NSCT_SR | SeAFusion | ResNet | RP_SR | HMSD_GF |
| 6 | IFCNN | MGFF | CNN | GFF | Hybrid_MSD | IFCNN | HMSD_GF | GFF |
| 7 | CNN | CNN | SwinFusion | HMSD_GF | TIF | TIF | NSCT_SR | RP_SR |
| 8 | HMSD_GF | HMSD_GF | Hybrid_MSD | FPDE | VSMWLS | Hybrid_MSD | MGFF | TIF |
| 9 | LP_SR | LP_SR | MGFF | IFCNN | IFEVIP | MGFF | U2Fusion | IFCNN |
| 10 | VSMWLS | VSMWLS | YDTR | MGFF | LP_SR | VSMWLS | TIF | MGFF |
| 11 | DLF | ResNet | SeAFusion | RP_SR | YDTR | GFF | GFCE | SwinFusion |
| 12 | ResNet | ADF | TIF | SwinFusion | ResNet | HMSD_GF | IFCNN | VSMWLS |
| 13 | FPDE | DLF | GFCE | GFCE | U2Fusion | CNN | ADF | ADF |
| 14 | ADF | FPDE | DLF | U2Fusion | LatLRR | GTF | CBF | SeAFusion |
| 15 | MSVD | RP_SR | LP_SR | VSMWLS | MGFF | LP_SR | FPDE | CBF |
| 16 | SeAFusion | MSVD | FPDE | IFEVIP | FPDE | RP_SR | VSMWLS | U2Fusion |
| 17 | SwinFusion | GTF | ADF | GTF | ADF | U2Fusion | LatLRR | FPDE |
| 18 | YDTR | SwinFusion | ResNet | LatLRR | GFCE | YDTR | SwinFusion | GFCE |
| 19 | RP_SR | SeAFusion | MSVD | SeAFusion | RP_SR | CBF | IFEVIP | IFEVIP |
| 20 | GFCE | YDTR | IFEVIP | TIF | DLF | GFCE | DLF | YDTR |
| 21 | LatLRR | NSCT_SR | RP_SR | DLF | GFF | IFEVIP | ResNet | DLF |
| 22 | IFEVIP | GFCE | GFF | YDTR | MSVD | LatLRR | SeAFusion | LatLRR |
| 23 | NSCT_SR | LatLRR | CBF | CBF | NSCT_SR | NSCT_SR | MSVD | GTF |
| 24 | CBF | CBF | NSCT_SR | ResNet | CBF | SeAFusion | YDTR | ResNet |
| 25 | GTF | IFEVIP | GTF | MSVD | GTF | SwinFusion | GTF | MSVD |

*Note: ranks follow descending T-BT strength; ties are broken alphabetically. Humans denotes the ranking derived from ground-truth labels. LPIFM–Humans MARD = 1.68; top-1 agreement: yes; top-3 overlap: 2/3. The best objective MARD in this table is SCD at 5.20; the worst is FMI at 6.88. The corresponding All-I/All-M rank table is given in Table S3.*

Table S8. Ablation on learning-rate schedulers and related hyperparameters.

| ID | Setting | Key change | Best Acc. (%) | *Δ* (pp) |
|---|---|---|---|---|
| **S1†** | **Cosine + warmup** | **warmup = 10, min. LR = 1×10$^{-5}$** | **78.78±0.87** | — |
| S2 | OneCycle | scheduler = OneCycle | 70.08 | -8.70 |
| S3 | ReduceLROnPlateau | scheduler = plateau | 79.00 | +0.22 |
| S4 | No warmup | warmup = 0 | 78.67 | -0.11 |
| S5 | Short warmup | warmup = 5 | 79.06 | +0.28 |
| S6 | Long warmup | warmup = 20 | 77.56 | -1.22 |
| S7 | Low start factor | start factor = 0.01 | 78.28 | -0.50 |
| S8 | High start factor | start factor = 0.3 | 77.23 | -1.55 |
| S9 | Low minimum LR | min. LR = 1×10$^{-6}$ | 63.55 | -15.23 |
| S10 | High minimum LR | min. LR = 3×10$^{-5}$ | 78.78 | 0.00 |

*Note: † the default configuration (five-run mean ± s.d.); other rows are single runs changing the stated setting. LR = learning rate. Δ is computed against the S1 mean.*

Table S9. Performance of 21 assessors on the scene-disjoint EVAFusion test split (170 scenes, 9,350 pairs), with exact equality in Overall score defining a Tie.

| **Rank** | **Assessor** | **Acc** | **macro-$F1$** | $\boldsymbol{\rho}$ |
|---|---|---|---|---|
| 1 | LPIFM fine-tuned | 0.5066 | 0.4941 | 0.4449 |
| 2 | Qcv | 0.4802 | 0.4099 | 0.3840 |
| 3 | Qabf | 0.4801 | 0.4069 | 0.3922 |
| 4 | FMI | 0.4509 | 0.3797 | 0.3170 |
| 5 | Mutinf | 0.4386 | 0.3733 | 0.2665 |
| 6 | MS-SSIM | 0.4299 | 0.3644 | 0.2520 |
| 7 | Ssim | 0.4290 | 0.3670 | 0.2422 |
| 8 | SCD | 0.4072 | 0.3419 | 0.1804 |
| 9 | CC | 0.3985 | 0.3393 | 0.1585 |
| 10 | VIFF | 0.3967 | 0.3346 | 0.1548 |
| 11 | Spatial_frequency | 0.3814 | 0.3211 | 0.1100 |
| 12 | Qcb | 0.3777 | 0.3213 | 0.0976 |
| 13 | Nabf | 0.3766 | 0.3172 | 0.0918 |
| 14 | Edge_intensity | 0.3745 | 0.3140 | 0.0840 |
| 15 | Avg_gradient | 0.3706 | 0.3115 | 0.0692 |
| 16 | LPIFM zero-shot | 0.3656 | 0.3265 | 0.0657 |
| 17 | Variance | 0.3632 | 0.3026 | 0.0520 |
| 18 | Entropy | 0.3626 | 0.3013 | 0.0461 |
| 19 | Psnr | 0.3266 | 0.2732 | -0.0479 |
| 20 | Rmse | 0.3256 | 0.2755 | -0.0503 |
| 21 | Cross_entropy | 0.3202 | 0.2772 | -0.0695 |

*Note: Metric names follow the VIFB naming convention [1].*

# Supplementary Analyses

## S1. Supplementary analysis of method-pool composition

The six held-out methods occupy All-I/All-M human ranks 1, 2, 4, 16, 23, and 25, with win rates ranging from 0.822 to 0.060 (Table 5 of the main text). Their pairwise comparisons therefore span an unusually broad quality range. Similar gains on this restricted pool occur for other conventional metrics: from U-I/All-M to U-I/U-M, Qcv accuracy rises from 0.604 to 0.720 and VIFF accuracy from 0.577 to 0.653. The higher scores on the held-out slice are thus associated with its method composition and are not unique to SCD.

For SCD, the median absolute difference between the scores assigned to the two candidates is approximately 0.36 on held-out pairs, compared with approximately 0.20 over all pairs. The resampling results show how strongly its accuracy varies with the composition of the method pool. Across 3,000 random six-method subsets, mean SCD accuracy is approximately 0.629, whereas accuracy on the actual held-out subset is approximately 0.803, near the 99th percentile of the resampled distribution. Subsets spanning at least 20 positions in the human ranking have mean accuracy of approximately 0.68, while dense mid-tier subsets average approximately 0.53. Consistently, the correlation between a subset's human win-rate span and its SCD accuracy is approximately 0.69.

The ranking analysis shows the same dependence at the method level. Spearman correlation between human win rates and mean SCD scores is 0.83 for the six held-out methods, 0.48 for the 19 training methods, and 0.67 over all 25 methods. This correlation is computed from human win rates and mean SCD scores and is distinct from the T-BT rank-correlation metric reported in the main evaluation tables. Taken together, the resampling, score-gap, and ranking results demonstrate that SCD is sensitive to method-pool composition: it performs particularly well for the widely separated methods in the held-out pool, while its accuracy is lower for pools dominated by closer mid-tier comparisons.

## S2. Additional consistency diagnostics

We conducted three post hoc diagnostics using the frozen SWA checkpoint, with the decision threshold $t = 0.3$ and calibration temperature $T_{cal} = 1.0$ fixed and without retraining or retuning. The primary analysis used U-I/All-M: 1,500 unordered pairs from five unseen scenes and the full pool of 25 methods. Because human labels were collected for unordered pairs, human candidate-swap consistency cannot be tested from these data.

For the candidate-swap diagnostic, every pair was scored in both orders. The hard criterion requires decisive verdicts to reverse and ties to remain ties; the soft preference gap must negate under the swap. LPIFM satisfied the hard criterion for all 1,500 pairs, with no violations among either decisive verdicts or ties. The Pearson correlation between the forward gap and the negative swapped gap was 1.0000. The mean absolute value of $d_{fwd} + d_{swp}$ was 0.0083, below $t = 0.3$, where $d_{fwd}$ and $d_{swp}$ denote the gaps for the forward and swapped candidate orders.

Cycle eligibility required three decisive edges within a method triplet. LPIFM produced 0 cycles among 10,054 eligible triplets. The human labels produced 70 cycles among 9,520 eligible triplets, corresponding to a pooled rate of 0.0074. Transitivity was evaluated over decisive two-edge paths, such as $A > B$ and $B > C$. The hard criterion requires a decisive closing verdict $A > C$, whereas the weak criterion also accepts a tie on the closing comparison. LPIFM achieved hard and weak transitivity of 1.0000 over 10,054 eligible paths. The corresponding human rates were 0.9624 and 0.9786 over 9,819 paths.

The secondary splits yielded the same pattern. Swap agreement was 100% on U-I/U-M, 100% on All-I/U-M, and 99.98% on All-I/All-M; the remaining 0.02% consisted only of tie-boundary cases. Soft swap correlation was 1.0000 throughout. LPIFM had zero cycles on every split, while pooled human cycle rates ranged from 0.0054 to 0.0110. LPIFM hard and weak transitivity were 1.0000 throughout, and human hard transitivity ranged from 0.9485 to 0.9840. The pre-specified high-consistency criteria were a swap-violation rate below 5%, a cycle rate no more than 2 pp (percentage points) above the human rate, and hard transitivity no more than 5 pp below the human rate. LPIFM met all three criteria. These diagnostics establish high internal consistency under the frozen settings; they do not establish agreement with human preferences, for which pairwise accuracy on U-I/All-M was 0.792.